\documentclass{article} % For LaTeX2e
\usepackage{iclr2027_conference,times}

\usepackage{amsmath,amsfonts,bm}

\def\eqref#1{equation~\ref{#1}}
\def\1{\bm{1}}

\DeclareMathAlphabet{\mathsfit}{\encodingdefault}{\sfdefault}{m}{sl}
\SetMathAlphabet{\mathsfit}{bold}{\encodingdefault}{\sfdefault}{bx}{n}

\usepackage{graphicx}
\usepackage{subcaption}   % preamble

\usepackage{hyperref}
\usepackage{url}
\usepackage{algorithm}
\usepackage{algpseudocode}
\algtext*{EndIf}% hide "end if" lines in algorithms
\usepackage{booktabs}
\usepackage{tablefootnote}
\usepackage{xurl}

\usepackage{multirow}
\usepackage[dvipsnames]{xcolor}
\usepackage{colortbl}

\usepackage{amsmath}

\newcommand{\runin}[1]{\par\vspace{0.75ex}\textbf{#1}\hspace{1em}\ignorespaces}
\newcommand{\gain}[1]{\textcolor{ForestGreen}{\scriptsize #1}}

\usepackage[nameinlink]{cleveref}
\crefname{appendix}{Appendix}{Appendices}\Crefname{appendix}{Appendix}{Appendices}

\title{Copy-Paste Language Models}

\author{Nathan Godey and Yoav Artzi\\
Cornell University\\
New York City, USA \\
\texttt{\{nathangodey,yoavartzi\}@cornell.edu} \\
}

\iclrfinalcopy % Uncomment for camera-ready version, but NOT for submission.
\begin{document}

\maketitle
\lhead{Pre-print.}

\begin{abstract}
Transformers are intuitively considered to be efficient at copying from their inputs. We actually show that the ability of LLMs to copy in context can be improved through explicit modeling. We propose Copy-Paste Language Models (CPLMs), which add a lightweight Copy-Paste head to the conventional language modeling (LM) head. CPLMs improve training efficiency, downstream performance, and offer new operating points in long-context and retrieval setups. We validate our approach by pretraining models from 135M to 1.7B parameters, showing improvements across all scales and backbones with negligible training and inference overheads. We explain the success of CPLMs by showing that they partially mitigate the LM head gradient bottleneck through an alternative gradient path. 
% We release code and checkpoints at
\end{abstract}
\section{Introduction}

% \ya{this is not a good start. I think you can motivate this in a more foundational way. many sequences see repetitions, natural language revisits events and entities, code and formal languages revisit variables, etc. current models learn to copy these through complex mechanism, but copying is a simple action to model directly, so why make our life hard, why make learning overly complex, this is such a simple and fundamental action, just create an avenue for the model to easily go through, and then leave it to learning to figure it out. }
% \gn{I get your point, but this only works with the bottleneck. Something very much like our copy mechanism could be happening e.g. on the final layer (usual attention to what you want to copy, then decode to corresponding proba). It would not be a much more complex mechanism than what we have. The complexity of learning this only comes from the limited power of the LM head.}

\begin{figure*}[b!]
  \centering
  % --- three panels in one row ---
  \begin{subfigure}[t]{0.55\linewidth}
    \includegraphics[width=\linewidth]{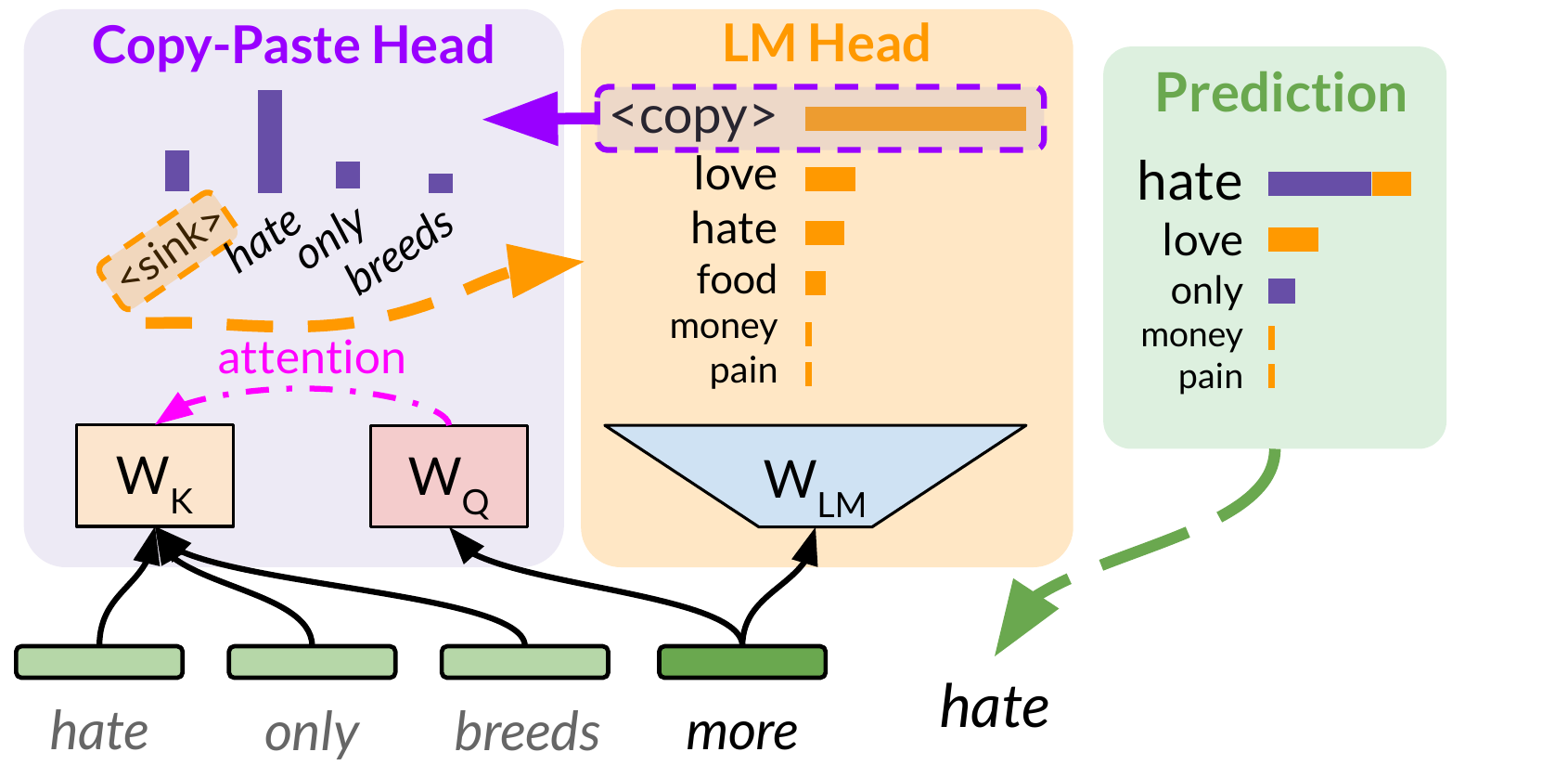}
    % \caption{}
  \end{subfigure}\hfill
  \begin{subfigure}[t]{0.45\linewidth}
    \includegraphics[width=\linewidth]{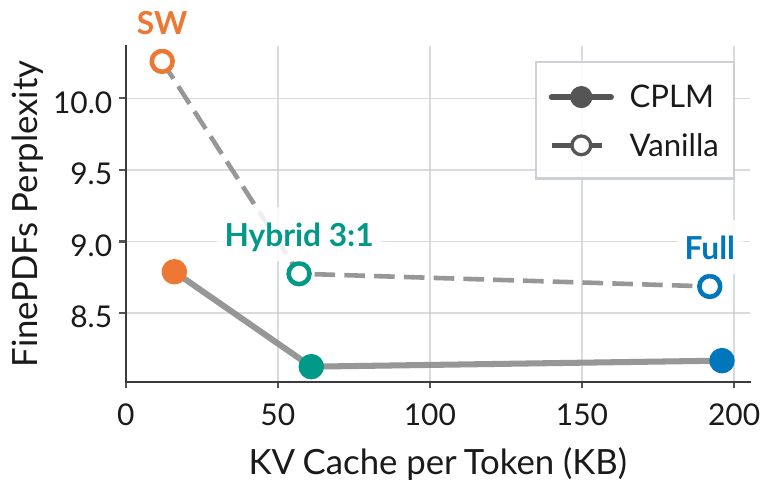}
    % \caption{FinePDFs PPL vs. per-token cache memory}
  \end{subfigure}
  \caption{
  % \textbf{Left:} Schema of our Copy-Paste (CP) head: The LM head predicts a \texttt{<copy>} token that gates a distribution on in-context tokens. A \texttt{<sink>} token triggers a rollback to the LM head prediction if no relevant token can be copied. The final prediction is a mixture of the LM \& CP heads' distributions. \textbf{Right:} Average FinePDFs perplexity for our 1.7B vanilla and CPLM models across different token-mixing architectures with different memory footprints. CPLMs outperform their vanilla counterparts by 1-1.5 PPL points, and the Sliding-Window CPLM matches the Full attention LM while using $\times$10 less memory per token.
  \textbf{Left:} Our Copy-Paste (CP) head. The LM head uses the probability of a special \texttt{<copy>} token to gate the copy distribution; a special \texttt{<sink>} token allows rolling back to the LM head. The prediction mixes both heads. \textbf{Right:} FinePDF perplexity of our 1.7B models vs. memory footprint. CPLMs outperform their vanilla counterparts by 1--1.5 PPL, and the sliding-window (SW) CPLM matches a full-attention LM at $12\times$ less memory per token, copy cache included.}

  \label{fig:teaser}
\end{figure*}

Language is rich with lexical repetitions. Documents revisit entities and events, dialogues echo earlier turns, and code refers back to the same variables and functions. As a result, $\sim$60\% of the tokens in FineWeb-Edu~\citep{fwedu}, a popular pre-training corpus, already appear earlier in their 4k-token context. Copying from the context is therefore one of the most basic operations a language model (LM) performs. 
Indeed, Transformer LMs devote many attention heads to copying, such as induction heads~\citep{olsson2022context} and retrieval heads~\citep{wu2025retrieval,lin-etal-2026-retrieval}.

Yet the standard neural LM setup can make copying surprisingly hard to learn. Every prediction is produced by the LM head, a linear projection far narrower than the vocabulary. This low-rank layer limits the distributions a model can express~\citep{softmax_bottleneck,finlayson2024closing}, while also discarding most of the training signal before it reaches the backbone~\citep{grad_bottleneck}. Copying is exposed to both effects, however easily the backbone can model the copy mechanism. As a consequence, vanilla Transformer models fail on a trivial synthetic copying task at large vocabulary sizes (\Cref{sec:analysis:softmax}). 
% Moreover, Transformer LMs devote numerous heads to induction~\citep{olsson2022context,wu2025retrieval,lin-etal-2026-retrieval}\gn{check new citation wu2025retrieval}. 
This raises the question of whether copying could be modeled more efficiently.

We introduce Copy-Paste Language Models (CPLMs), which are designed to facilitate copying in LMs. CPLMs are inspired by pointer networks~\citep{pointer_vinyals}, which were designed to give encoder-decoder and recurrent models direct access to their inputs, and largely fell out of favor with the emergence of self-attention. 
Yet we argue that a key obstacle to copying lies on the output side, and that a pointer offers a way around the LM head bottleneck. A CPLM adds a Copy-Paste (CP) head adjacent to the LM head, which predicts the next token by pointing at a past position in the context. The LM head decides when to copy by emitting a dedicated \texttt{<copy>} token, and a sink lets the CP head defer back to the LM head when nothing in the context is to be copied. 

This copy mechanism is learned directly in the vocabulary space, and adds a second, much less compressed gradient pathway next to the LM head. Our contribution is thus not the adaptation of the copy mechanism itself, which is related to prior pointer models, but showing that such a head substantially improves modern LM pretraining across backbones, and that it does so mostly by mitigating the LM head gradient bottleneck, as supported by targeted interventions (\Cref{sec:analysis:softmax}). The CP head adds two lightweight projection matrices and negligible parameter and compute overheads (resp. $0.5 \%$ and $3 \%$ at 1.7B scale, and below $1\%$ of training throughput at 9B and 30B; \Cref{app:efficiency}).

Making copying easy to learn yields better models overall. We pretrain CPLMs from 135M to 1.7B parameters on up to 200B tokens across full-attention, sliding-window, hybrid and Gated DeltaNet backbones. CPLMs converge faster, reaching the final loss of their vanilla counterparts up to $2\times$ faster, and improve perplexity at every scale and on every backbone, by 0.5--1.5 points on out-of-distribution FinePDFs documents at 1.7B scale. These gains are not restricted to copied tokens: CPLMs also better predict the tokens they do not copy (\Cref{app:additional:copy_nll}). Critically, the benefits carry over to supervised fine-tuning, where CPLMs improve on math, code and instruction following by 1--5 average accuracy points.

CPLMs also change how models use their context. Sliding-window CPLMs match full-attention and hybrid models using up to $12\times$ less memory per token, and they retrieve information from contexts far longer than those seen in training much better than their vanilla counterparts. CPLMs also rely on fewer induction heads in their backbone, suggesting that the CP head takes over part of their role. Moreover, the CP head can read documents through their copy keys alone, which offers a memory-efficient alternative for retrieval-augmented generation (RAG) setups.

Finally, CPLMs demonstrate significant benefits even in extremely optimized environments:  CPLMs improve on the prior NanoGPT speedrun~\citep{modded_nanogpt_2024} record by 11.3\%, achieving the largest architecture-centric speed-up -- and second largest overall -- since December 2024.\footnote{\url{https://github.com/KellerJordan/modded-nanogpt/pull/379}}

% Many recent advances in language model (LM) pretraining have been driven by architectural improvements for the hidden layers \citep{deepseek2,gated_dn,kimiteam2026kimik3openfrontier}, new optimizers \citep{shampoo, muon_scalable}, or novel training objectives \citep{token_super,mtp}. However, the process of token prediction itself remains largely unmodified, relying on a simple linear projection followed by a softmax normalization, often called the \textit{LM head}.

% In this work, we propose to augment the token prediction process of neural language models with a \textit{Copy-Paste head}. This Copy-Paste (CP) head learns to copy tokens directly from context, working as a supplement to the traditional LM head. This lightweight module addition substantially improves language modeling performance across all tested sizes and setups.

% We validate the efficacy of the CP head by training 1.7B parameter models on 200B tokens

\section{Method}

\subsection{Background \& Notation}

\runin{Standard LM Head}\label{sec:meth:standard}
A neural language model with hidden dimension $D$ processes a token sequence $x_1, \dots, x_T$, taken from a vocabulary $\mathcal{V}$ of size $V$, and produces an output representation $h_t \in \mathbb{R}^D$ for each position $t$, which only depends on the prefix $x_{\le t}$. The LM head predicts the next token from $h_t$, using a learnable embedding $w_v \in \mathbb{R}^D$ for each token $v \in \mathcal{V}$. For scores $z_u$ indexed by a set $\mathcal{S}$, we write $\operatorname{softmax}_{u \in \mathcal{S}}(z_u)_v = \exp(z_v) / \sum_{u \in \mathcal{S}} \exp(z_u)$ for the probability $p(v \mid \cdot)$, with $v \in \mathcal{S}$. 
With the embeddings $w_v$ forming the LM head matrix $W \in \mathbb{R}^{V \times D}$, the LM distribution is:
\begin{equation}\label{eq:lm}
    p_{\mathrm{LM}}(v \mid x_{\le t}) \ = \ \operatorname*{softmax}_{u \in \mathcal{V}}\left(w_u^\top h_t\right)_v.
\end{equation}
% The embeddings $w_v$ form the LM head matrix $W \in \mathbb{R}^{V \times D}$.
%and the model is trained to maximize $\log p(x_{t+1} \mid x_{\le t})$, where $p$ denotes its final next-token distribution.

\runin{Softmax Bottleneck(s)}\label{sec:meth:softmax_bttl}
Commonly, the vocabulary size $V$ is much larger than the hidden dimension $D$. As all the logits $w_v^\top h_t$ are obtained from a $D$-dimensional vector, the LM head cannot represent every distribution over $\mathcal{V}$, which limits the model expressivity. This is commonly known as the softmax bottleneck, and has been shown to partly impede expressivity in practice~\citep{softmax_bottleneck,grivas-etal-2022-low,finlayson2024closing,basri2026the}. \citet{grad_bottleneck} observe that the LM head also bottlenecks optimization by destructively compressing the $V$-dimensional logit gradients to a $D$-dimensional output space during the backward pass. Our method reduces both these issues with a second prediction and gradient pathway that does not suffer
from such bottlenecks as much.

\subsection{Copy-Paste Language Models}

\runin{The Copy-Paste Head}\label{sec:meth:cp_head}
Two assumptions are crucial for the softmax bottleneck to happen: the target space is a relatively large token vocabulary, and the LM head is a fixed parameter across generation. We propose to alleviate these constraints by building a second pathway for next-token prediction that does not have a fixed target space, and where the prediction is not anchored to a static weight matrix.

The second head's targets are the context positions $\tau \in \{1, \dots, t\}$. The head scores each position against the current state, as in self-attention, using a query $q_t = W_Q h_t$ and keys $k_\tau = W_K h_\tau$:
\begin{equation}\label{eq:attn}
    \alpha_t(\tau) \ = \ \operatorname*{softmax}_{\tau' \in \{1, \dots, t\}}\left(k_{\tau'}^\top q_t / \sqrt{D}\right)_\tau\;\;.
\end{equation}
For simplicity, we set the dimension of the queries and keys to the hidden dimension $D$ by default, i.e. $W_Q, W_K \in \mathbb{R}^{D \times D}$; much smaller dimensions perform similarly (\Cref{app:ablations:copy_dim}).
We obtain a distribution over tokens by marginalizing over the positions that emit the same token:
\begin{equation}\label{eq:cp}
    p_{\mathrm{CP}}(v \mid x_{\le t}) \ = \ \sum_{\tau \le t \,:\, x_\tau = v} \alpha_t(\tau)\;\;.
\end{equation}
This distribution is only supported on the tokens that appear in the context. Informally, this head computes a probability distribution over past positions, and pulls the corresponding token distribution to make a prediction. Because this intuitively corresponds to a copy and paste operation, we call this head a \emph{Copy-Paste} head (or CP head).

\runin{Triggering the CP head}
We add a dummy \texttt{<copy>} token to the LM head, with embedding $w_{\texttt{<copy>}}$, and use the probability that the LM head assigns to it as a gate for the copy-paste mechanism:
\begin{equation}\label{eq:gate}
    c_t \ = \ \operatorname*{softmax}_{u \in \mathcal{V} \cup \{\texttt{<copy>}\}}\left(w_u^\top h_t\right)_{\texttt{<copy>}}\;\;.
\end{equation}
Because \texttt{<copy>} is never an actual target and only serves as a gate, trained through $c_t$ in the mixture below, $p_{\mathrm{LM}}$ remains the distribution of \Cref{eq:lm} over the actual vocabulary, which is the LM head distribution without \texttt{<copy>}. The next-token distribution is then the mixture
\begin{equation}\label{eq:proba_no_sink}
    p(v \mid x_{\le t}) \ = \ (1 - c_t) \, p_{\mathrm{LM}}(v \mid x_{\le t})
                      \ + \ c_t \, p_{\mathrm{CP}}(v \mid x_{\le t})\;\;.
\end{equation}

\runin{Sink mechanism}
When the copy-paste mechanism is triggered, the support of the next-token distribution
can only include in-context tokens. This could induce failure modes where the model
attempts to copy a token that is not actually in the context, with no valid alternatives
available, resulting in a wrong prediction. To avoid this, we add a sink
mechanism to roll back to the LM head prediction. We add a
learnable sink key $k_{\texttt{<sink>}}$ to the keys of \Cref{eq:attn}, and write
\begin{equation}\label{eq:attn_sink}
    s_t \ = \ \operatorname*{softmax}_{\tau \in \{\texttt{<sink>}, 1, \dots, t\}}\left(k_\tau^\top q_t / \sqrt{D}\right)_{\texttt{<sink>}}
\end{equation}
for the probability that the CP head places on the sink rather than on any context position. $\alpha_t$ and $p_{\mathrm{CP}}$ remain defined over the context positions only (\Cref{eq:attn,eq:cp}). \Cref{eq:proba_no_sink} becomes
\begin{equation}\label{eq:proba_sink}
    \boxed{\ p(v \mid x_{\le t}) \ = \ (1 - \lambda_t) \, p_{\mathrm{LM}}(v \mid x_{\le t})
                             \ + \ \lambda_t \, p_{\mathrm{CP}}(v \mid x_{\le t}),
    \qquad \lambda_t \ = \ c_t \, (1 - s_t)\;\;. \ }
\end{equation}
The effective gate $\lambda_t$ is the probability of routing through the CP head
\emph{and} not falling back to the sink: $c_t$ opens the copy pathway, and $s_t$ lets the model close it
again when no context position is a suitable target. Setting $s_t = 0$ recovers
\Cref{eq:proba_no_sink}. As a convex combination of two distributions over
$\mathcal{V}$, \Cref{eq:proba_sink} is a valid distribution by construction.
% , and the two
% pathways receive gradient in proportion to $1 - \lambda_t$ and $\lambda_t$ respectively
CPLMs are trained end-to-end with the standard next-token cross-entropy on $p$ only: there is no auxiliary loss and no copy supervision, and the model learns on its own when and where to copy.
\Cref{eq:proba_sink} is close to the pointer sentinel mixture of \citet{w103}, where a sentinel inside the pointer softmax is the only gate: its probability weights the vocabulary distribution. In CPLMs, the gate $c_t$ is instead predicted by the LM head together with the vocabulary, so that the copy decision is known before any copy attention is computed, and the sink only lets the CP head defer back to the LM head.

\section{Experiments \& Results}\label{sec:results}

\subsection{Pretraining}\label{sec:results:pt}

\begin{figure*}[t]
  \centering
  % --- shared legend, centered on top ---
  \includegraphics[height=1.5em]{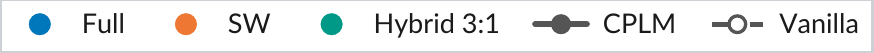}\\
  % --- four panels in one row ---
  \begin{subfigure}[t]{0.33\linewidth}
    \includegraphics[width=\linewidth]{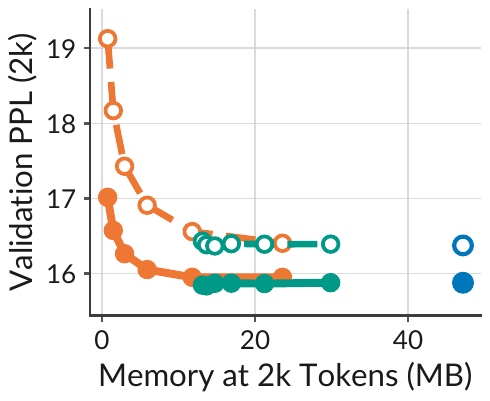}
    \caption{FineWeb-Edu}
  \end{subfigure}\hfill
  \begin{subfigure}[t]{0.33\linewidth}
    \includegraphics[width=\linewidth]{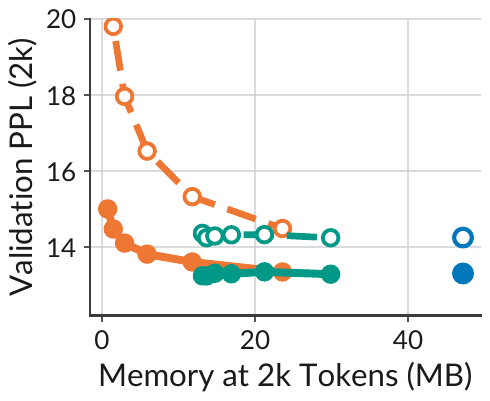}
    \caption{FinePDFs (OOD)}
  \end{subfigure}\hfill
  \begin{subfigure}[t]{0.33\linewidth}
    \includegraphics[width=\linewidth]{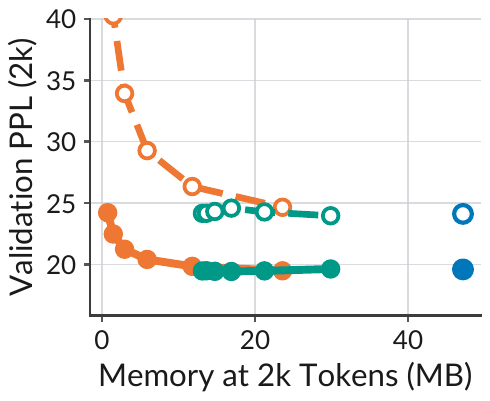}
    \caption{arXiv (OOD)}
  \end{subfigure}\hfill
  % \begin{subfigure}[t]{0.245\linewidth}
  %   \includegraphics[width=\linewidth]{imgs/ppl_vs_memory_norm_govreport.pdf}
  %   \caption{GovReport}
  % \end{subfigure}
  \caption{Validation perplexity vs. memory usage for sliding window sizes from 32 to 1024 at 135M scale. CPLMs improve the memory--performance frontier.}
  \label{fig:ppl_vs_memory}
\end{figure*}

We pretrain models ranging from 135M to 1.7B parameters on up to 200B tokens of FineWeb-Edu data \citep{fwedu}. To study the interaction between backbone context handling and the CP head, we explore various backbones: full attention (Full), sliding-window (SW), and a hybrid architecture (Hybrid) where full-attention and sliding-window layers are interleaved. Hybrid models give sliding-window backbones access to the whole context through a few global layers, and thus serve as a natural baseline for adding global access to local backbones, as for Sliding-Window CPLMs. Full and Hybrid backbones already access every position, so that CPLM gains on these backbones cannot stem from adding global access alone. We also report Gated DeltaNet~\citep[GDN; ][]{gated_dn} and Hybrid-GDN comparisons in \Cref{app:ablations:gdn}. At 135M scale, we run several ablations, notably varying the backbone sliding window $w$ for SW and Hybrid models, from 32 to 1{,}024. Based on our findings (\Cref{fig:ppl_vs_memory}), we set $w=256$ for all Hybrid and SW 1.7B models. Our 61 135M-scale runs cost ${\sim}50$ GH200 hours each, and our 1.7B runs ${\sim}900$ B200 hours each. We follow the SmolLM2 recipe~\citep{allal2025smollm2smolgoesbig}, with identical hyperparameters for both arms (\Cref{app:hyperparams}). All checkpoints are available on HuggingFace~\footnote{\url{https://huggingface.co/collections/nthngdy/cplm}}.

\begin{figure*}[t]
  \centering
  % --- shared legend, centered on top ---
  \includegraphics[height=1.5em]{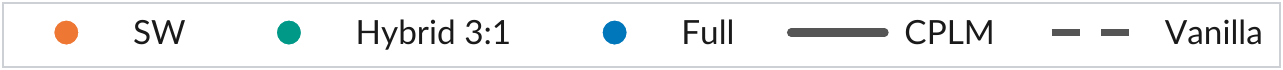}\\
  % --- three panels in one row ---
  \begin{subfigure}[t]{0.41\linewidth}
    \includegraphics[width=\linewidth]{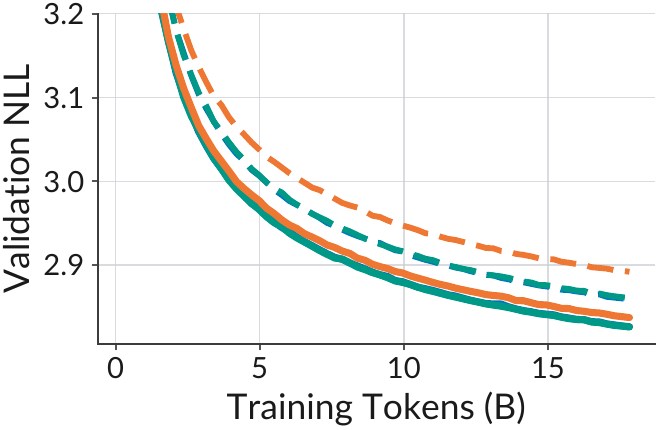}
    \caption{135M ($\sim$20B tokens)}
  \end{subfigure}\hfill
  % \begin{subfigure}[t]{0.32\linewidth}
  %   \includegraphics[width=\linewidth]{imgs/val_nll_curves_360M.pdf}
  %   \caption{360M ($\sim$20B tokens)}
  % \end{subfigure}\hfill
  \begin{subfigure}[t]{0.41\linewidth}
    \includegraphics[width=\linewidth]{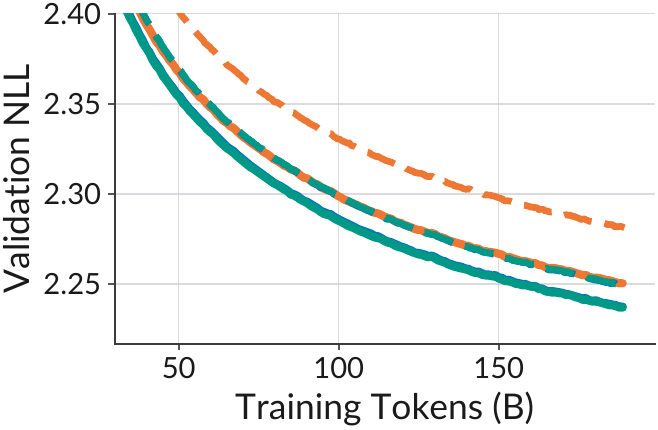}
    \caption{1.7B ($\sim$200B tokens)}
  \end{subfigure}
  \caption{FineWeb-Edu validation loss vs. training tokens (before LR decay), for $w=256$. Colors distinguish backbone attention setups. Full and Hybrid training curves are essentially superposed.}
  \label{fig:val_nll_curves}
\end{figure*}

We report our training dynamics for CPLMs and vanilla LMs at 135M and 1.7B scale in \Cref{fig:val_nll_curves}, and at 360M scale in \Cref{app:ablations:360m}. The dynamics are favorable to CPLMs in all configurations, reaching the final loss of vanilla models $1.25\times$ faster in the Full and Hybrid attention setups, and $2\times$ faster in the Sliding Window (SW) setup at 1.7B scale. The 1.7B SW CPLM matches the Full LM with $12\times$ less memory at 4k tokens, as it only caches one copy key per token beyond its window.

\begin{table}[t]
  \centering
  \resizebox{\linewidth}{!}{%
  \begin{tabular}{llccccccc|c}
    \toprule
    \textbf{Model} & \textbf{Type} & \textbf{HellaSwag} & \textbf{LAMBADA} & \textbf{PIQA} & \textbf{SciQ} & \textbf{ARC-e} & \textbf{ARC-c} & \textbf{WinoGrande} & \textbf{Avg.} \\
    \midrule
    Full & Vanilla & 61.12 & 44.46 & \textbf{75.24} & 88.20 & \textbf{68.39} & 40.96 & 58.80 & 62.45 \\
     & CPLM & \textbf{61.70} & \textbf{46.92} & 74.86 & \textbf{89.20} & 68.10 & \textbf{41.21} & \textbf{59.43} & \textbf{63.06} \\
    \midrule
    SW & Vanilla & 61.00 & 42.46 & 74.48 & 87.00 & 67.17 & \textbf{41.30} & 58.01 & 61.63 \\
     & CPLM & \textbf{61.78} & \textbf{46.42} & \textbf{74.70} & \textbf{87.30} & \textbf{67.59} & 41.13 & \textbf{58.41} & \textbf{62.48} \\
    \midrule
    Hybrid & Vanilla & 61.08 & 43.70 & 74.48 & \textbf{87.80} & \textbf{69.32} & 40.96 & 58.17 & 62.22 \\
     & CPLM & \textbf{61.70} & \textbf{46.92} & 74.48 & 87.40 & 66.46 & \textbf{41.55} & \textbf{59.59} & \textbf{62.59} \\
    \bottomrule
  \end{tabular}}
  \caption{Zero-shot accuracy (\%) of our 1.7B base models (95k steps). Best of each pair in bold.}
  \label{tab:nlu-0shot}
\end{table}

\runin{Downstream Evaluations}In \Cref{tab:nlu-0shot}, we report the zero-shot performance of the 1.7B models on standard benchmarks: HellaSwag~\citep{zellers2019hellaswag}, LAMBADA~\citep{paperno2016lambada},
PIQA~\citep{piqa}, SciQ~\citep{sciq},
ARC~\citep{allenai:arc}, and
WinoGrande~\citep{winogrande}. Overall, CPLMs outperform their vanilla counterparts by +0.4--0.9 accuracy points while inducing no meaningful compute overhead ($\sim$3\%, \Cref{app:efficiency}). The largest gains are on LAMBADA (+2.5--4.0 points), whose target word often appears earlier in the passage and can thus be copied, while gains on the other tasks are smaller. However, the gains of CPLMs are not restricted to copying. Token-level NLL decreases across all tokens, including when the \texttt{<copy>} token receives near-zero probabilities (\Cref{app:additional:copy_nll}).

\runin{Effect of the Window Size $w$}\Cref{fig:ppl_vs_memory} shows the impact of the sliding window parameter $w$ across different document sources, allowing direct comparison between CPLMs and vanilla LMs. Across all window sizes $w$, CPLMs have lower perplexity, dropping by 0.5--2 points on in-domain FineWeb-Edu, and by up to 15 points on other datasets, a trend that extends to non-English data (\Cref{app:analysis:lang}). Moreover, a SW CPLM with a relatively small window ($w\simeq 128$) generally matches the perplexity of a full-attention vanilla model while using much less memory. Comparing SW CPLMs and vanilla Hybrid models is particularly meaningful, as in both cases the architectures mix sliding-window layers and layers where the full context can be accessed directly (via full-attention or the CP head). SW CPLMs outperform all window-matched vanilla Hybrid, which shows that the CP head lets the model make better use of context compared to interleaved full-attention layers, while using less memory.

\begin{figure*}[t]
  \centering
  % --- shared legend, centered on top ---
  \includegraphics[height=1.5em]{imgs/val_nll_curves_legend.pdf}\\
  % --- three panels in one row ---
  \begin{subfigure}[t]{0.32\linewidth}
    \includegraphics[width=\linewidth]{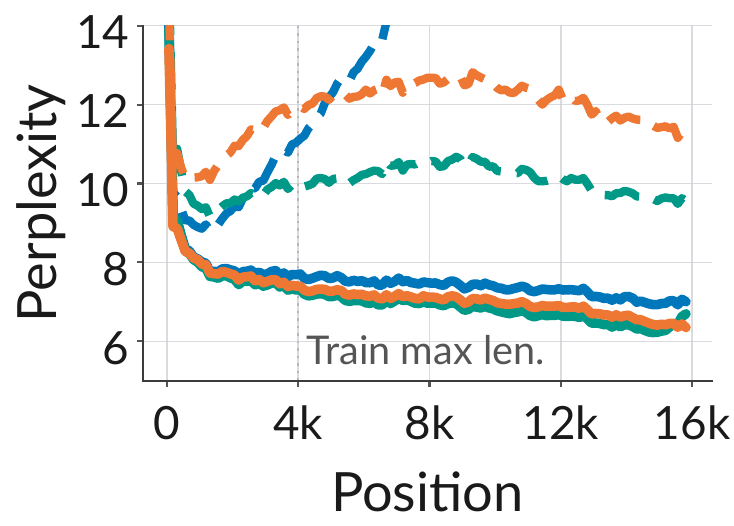}
    \caption{Perplexity (OOD)}\label{fig:rolling:ppl}
  \end{subfigure}\hfill
  \begin{subfigure}[t]{0.32\linewidth}
    \includegraphics[width=\linewidth]{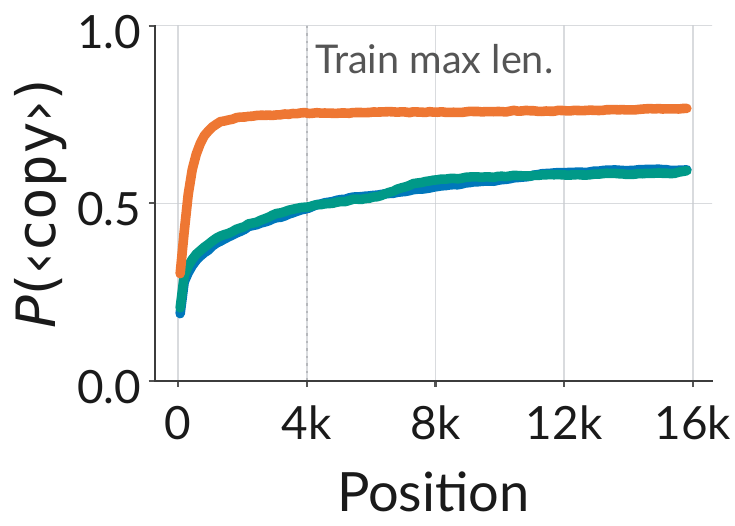}
    \caption{Copy gate}\label{fig:rolling:gate}
  \end{subfigure}\hfill
  \begin{subfigure}[t]{0.32\linewidth}
    \includegraphics[width=\linewidth]{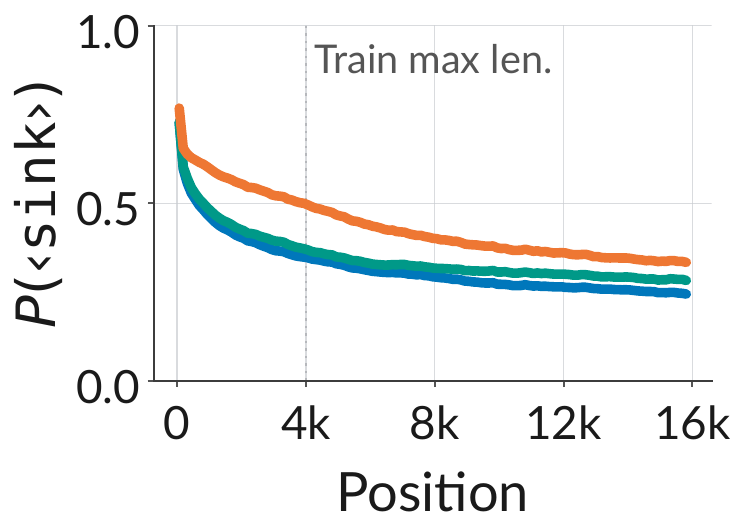}
    \caption{Sink}\label{fig:rolling:sink}
  \end{subfigure}
  \caption{Per-position statistics of our 1.7B models on 5{,}000 FinePDFs documents of 16k tokens, i.e. up to $4\times$ the pretraining context. Full and Hybrid models use YaRN ($\times$4). \textbf{(a)} Perplexity. \textbf{(b--c)} Average probability of the \texttt{<copy>} token and of the sink. CPLMs keep improving with context beyond their training length, and use the CP head more as the context grows.
  }
  \label{fig:rolling}
\end{figure*}

\runin{Rolling Perplexity}\Cref{fig:rolling} reports the perplexity of our 1.7B models as a function of the position in the document, up to $4\times$ their pretraining context. The perplexity of CPLMs decreases steadily with position across all backbones, including beyond the training length, reaching 6.5--7 at 16k tokens. In contrast, the perplexity of all vanilla models increases after roughly 1k tokens: SW and Hybrid models plateau around 10--13, and the Full model diverges beyond its training length. The CP head is used more as the context grows: the \texttt{<copy>} probability increases with position while the sink probability decreases, and the average copy distance grows almost linearly with position (\Cref{app:analysis:copy_dist}). The SW CPLM relies on the CP head the most, and copies tokens more than 2,000 positions back at 16k tokens, far beyond its window of 256 tokens.

\runin{Ablations} We conduct ablations at 135M scale to explore and validate several aspects of CPLMs (\Cref{app:ablations}). We run a multi-seed analysis at 135M scale, showing that our gains are 15--40$\times$ larger than the seed-to-seed standard deviation ($p < 10^{-3}$), in line with the low sensitivity of pretraining loss to random seeds~\citep{vanderwal2025polypythias}. We show that the dimension of the queries and keys in the CP head barely matters, with dimension $64$ achieving a similar performance to dimension $2{,}048$, and that the sink matters most for small windows (\Cref{app:ablations:sink}). Finally, we compare the CP head to alternative pointer designs (\Cref{sec:output-ablations}): it outperforms the partitions that \citet{chang-etal-2023-revisiting-architectures} proposed for decoder LMs, and matches or outperforms sigmoid-gated~\citep{see-etal-2017-get} and shared-softmax~\citep{gu-etal-2016-incorporating} mixtures by small margins.

\subsection{Supervised Fine-Tuning (SFT)}
We hypothesize that stronger token copying abilities result in better chain-of-thought reasoning and coding abilities, because copy is inherently important for these tasks. To validate this hypothesis, we perform Supervised Fine-Tuning (SFT) on our pretrained 1.7B models. We use the Tulu-3 dataset~\citep{lambert2025tulu} with sequences of up to 16k tokens. 

\begin{table}[t]
  \centering 
  \setlength{\tabcolsep}{5pt}
  \resizebox{\linewidth}{!}{%
  \begin{tabular}{llcccccccccccc}
  \toprule
   & & \multicolumn{2}{c}{PPL ($\downarrow$)} & & \multicolumn{2}{c}{Math} & & \multicolumn{2}{c}{Coding} & & Instruction & & \\
  \cmidrule(lr){3-4} \cmidrule(lr){6-7} \cmidrule(lr){9-10} \cmidrule(lr){12-12}
  Attention & Model & Tulu-3 & Nemotron-3 & & MATH & GSM8K & & HumanEval & MBPP & & IFEval & & Avg \\
  \midrule
  Full & Vanilla & 2.52 & 4.08 & & 13.9 & 28.9 & & 20.4 & 16.9 & & 41.2 & & 24.2 \\
   & CPLM & \textbf{2.47} & \textbf{3.78} & & \textbf{17.5} & \textbf{33.1} & & \textbf{21.1} & \textbf{17.5} & & \textbf{43.0} & & \textbf{26.4} \\
  \midrule
  Hybrid 3:1 & Vanilla & 2.52 & 4.10 & & 16.5 & 27.9 & & 19.8 & \textbf{19.0} & & 38.3 & & 24.3 \\
   & CPLM & \textbf{2.48} & \textbf{3.80} & & \textbf{17.4} & \textbf{30.4} & & \textbf{22.1} & 17.4 & & \textbf{40.1} & & \textbf{25.5} \\
  \midrule
  SW-256 & Vanilla & 2.79 & 5.65 & & 7.2 & 18.6 & & 15.4 & 16.1 & & 37.2 & & 18.9 \\
   & CPLM & \textbf{2.54} & \textbf{4.08} & & \textbf{16.4} & \textbf{25.2} & & \textbf{19.2} & \textbf{20.8} & & \textbf{38.4} & & \textbf{24.0} \\
  \bottomrule
  \end{tabular}}
  \caption{Downstream results after Supervised Fine-Tuning on our 1.7B models. PPL: Tulu-3 val. and \texttt{Nemotron3-SFT-Math-v4} perplexity. Evaluation details in \Cref{app:sft_evals}.}
  \label{tab:math-cot}
\end{table}

\Cref{tab:math-cot} reports evaluation metrics for SFT 1.7B models. CPLMs outperform Vanilla LMs by 1.2--5.1 points in average score, and on every benchmark except MBPP for the Hybrid model. The Sliding Window CPLM is once again on par with full-attention models while using significantly less memory. The gap is particularly consistent on math benchmarks that benefit from chain-of-thought reasoning, and holds for every $k$ in pass@$k$ (\Cref{fig:gsm8k_passk}). For mathematical tasks, CPLMs outperform their Vanilla baselines by +3.9 average points for full-attention models, and +7.9 for sliding window models. This downstream performance improvement is accompanied by the lower perplexity of CPLMs on both Tulu-3 and Nemotron-3 SFT data~\citep{nvidia2025nvidianemotron3efficient}, hinting that the CP head improves modeling for reasoning data.

\subsection{Long-Context Abilities}

\begin{table}[t]
  \centering
  \resizebox{\columnwidth}{!}{%
  \begin{tabular}{llccccccccccc}
  \toprule
   & & & \multicolumn{2}{c}{Synthetic (0--16k)} & & \multicolumn{7}{c}{LongBench (16k)} \\
  \cmidrule(lr){4-5} \cmidrule(lr){7-13}
  Attention & Model & & BABILong & RULER & & Single-Doc & Multi-Doc & Summ. & Few-shot & Synth. & Code & \textbf{Avg} \\
  \midrule
  Full & Vanilla & & 26.4 & 33.6 & & 5.7 & 6.1 & \textbf{11.8} & 20.7 & \textbf{1.7} & 15.4 & 10.4 \\
   & CPLM & & \textbf{32.0} & \textbf{39.6} & & \textbf{6.8} & \textbf{6.7} & 10.2 & \textbf{28.3} & 0.8 & \textbf{18.6} & \textbf{12.2} \\
  \midrule
  Hybrid 3:1 & Vanilla & & 27.7 & 33.2 & & 4.6 & \textbf{5.3} & 8.4 & 20.5 & 0.7 & 17.4 & 9.5 \\
   & CPLM & & \textbf{32.9} & \textbf{37.8} & & \textbf{5.4} & 3.3 & \textbf{14.0} & \textbf{35.1} & \textbf{1.7} & \textbf{25.6} & \textbf{14.3} \\
  \midrule
  SW-256 & Vanilla & & 13.8 & 7.9 & & 3.0 & 2.9 & 6.4 & 16.6 & 1.4 & 14.8 & 7.4 \\
   & CPLM & & \textbf{16.1} & \textbf{29.2} & & \textbf{7.7} & \textbf{4.4} & \textbf{10.8} & \textbf{30.3} & \textbf{2.1} & \textbf{25.0} & \textbf{13.4} \\
  \bottomrule
  \end{tabular}}
  \caption{Long-context results for the 1.7B base models. BABILong: mean
    accuracy over qa1--qa5 and lengths 0k--16k; RULER: mean over its 13 tasks
    and lengths 4k--16k; LongBench: category means at 16k context, Avg over the 16 tasks. Full and Hybrid use YaRN ($\times$4 from 4096); SW-256 uses plain
    RoPE. \textbf{Bold} marks the better model of each pair. Details in \Cref{app:long_ctx}.}
  \label{tab:longctx}
\end{table}

A model that copies better is likely to also retrieve information more accurately from long documents. We verify this in \Cref{tab:longctx}, where we show that our base 1.7B CPLMs outperform their vanilla counterparts on both synthetic tasks~\citep{hsieh2024ruler,babilong} by +7.5 average points, and on LongBench~\citep{bai-etal-2024-longbench} by +4.1 average points.

We also run a single-needle retrieval test. A 7-digit ``magic number''
is placed at one of 8 depths inside neutral filler text, for context lengths from 1k to 64k tokens, i.e. up to $16\times$ the pretraining context (details in \Cref{app:niah}). Averaged over the whole depth/length grid, CPLM outperforms Vanilla for every architecture. The gain is small for Full attention (49.7 $\rightarrow$ 54.8), large for Hybrid 3:1 (58.2 $\rightarrow$ 83.6), and very large for SW
(14.3 $\rightarrow$ 99.4). The SW CPLM finds the needle almost perfectly at every length and depth, with a window of only $w=256$, showing that the CP head is robust to out-of-distribution context lengths, and can retrieve well beyond the sliding window.

\subsection{Efficient Retrieval}\label{sec:res:hotpot}

The CP head can be considered a form of external memory. We study whether CPLMs can learn to copy from documents when the backbone has no access to the document's content.
We fine-tune the base 1.7B vanilla models and CPLMs
(Full attention, SW, and Hybrid 3:1) on the train set of HotpotQA~\citep{yang-etal-2018-hotpotqa} in the distractor setting.
To enable robustness to a variable number of distractor documents, each training sample contains the two gold paragraphs and $n \sim \mathcal{U}[5,50]$ distractor paragraphs drawn from a shared batch-level pool, with no context length truncation. All models are trained for one epoch, using a learning rate of $10^{-4}$ with cosine decay, and batch size 32. 

We  evaluate 200 validation questions with $n$ ranging from 5 to 500
distractors.
We experiment with two regimes: (a) \emph{docs visible}: the backbone attends over the whole input; and (b) \emph{docs dropped}: each document is encoded independently by the backbone, so the question and answer tokens cannot attend to the documents in the backbone layers, but only reach them through the CP head. As a control, we fine-tune vanilla baselines that we allow to reach the documents only through the last-layer keys and values. All models are fine-tuned in their evaluation regime.

\begin{figure*}[t]
  \centering
  % --- shared legend, centered on top ---
  \includegraphics[height=1.5em]{imgs/hotpot_yarn4_distractor_legend.pdf}\\
  % --- four panels in one row ---
  \begin{subfigure}[t]{0.33\linewidth}
    \includegraphics[width=\linewidth]{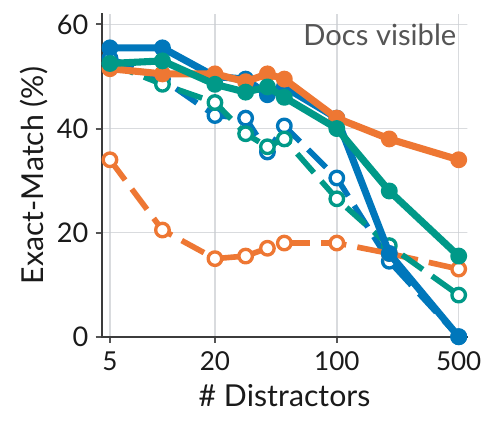}
    \caption{Document KV}\label{fig:hotpot_dist:u550}
  \end{subfigure}\hfill
  \begin{subfigure}[t]{0.33\linewidth}
    \includegraphics[width=\linewidth]{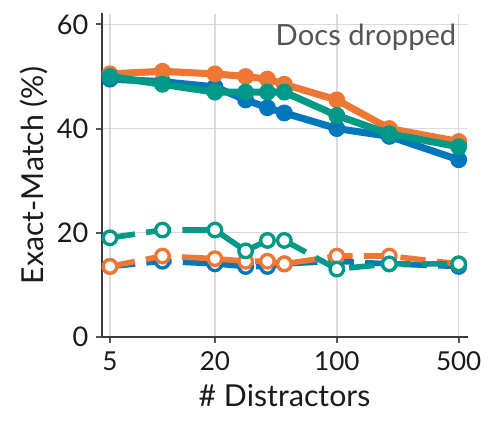}
    \caption{No Document KV}\label{fig:hotpot_dist:dropdocs}
  \end{subfigure}\hfill
  \begin{subfigure}[t]{0.33\linewidth}
    \includegraphics[width=\linewidth]{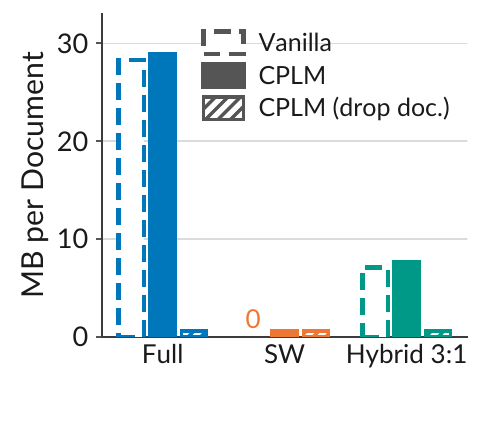}
    \caption{Memory requirements}\label{fig:hotpot_dist:mem}
  \end{subfigure}\hfill
  % \begin{subfigure}[t]{0.245\linewidth}
  %   \includegraphics[width=\linewidth]{imgs/ppl_vs_memory_norm_govreport.pdf}
  %   \caption{GovReport}
  % \end{subfigure}
  \caption{HotpotQA performance for 1.7B models. Models are fine-tuned and evaluated either with full document KV visible (\textbf{left}), or with only a single access to the document memory: last-layer KV for Vanilla, and CP states for CPLMs (\textbf{middle}).}
  \label{fig:hotpot_dist}
\end{figure*}

CPLMs match or outperform their vanilla counterparts across backbones and distractor counts (\Cref{fig:hotpot_dist:u550}) at a small memory overhead: the CP head keys add 2\% to the cache of full attention and 8\% to Hybrid 3:1 (\Cref{fig:hotpot_dist:mem}), and slow down decoding by less than 2\% at this scale (\Cref{app:efficiency}). The gap is largest with SW models, where the vanilla LM loses the documents that leave its window (15\% EM at 20 distractors) while the SW CPLM still reaches 34\% at 500 distractors, where every Full model fails. The CP head cache alone carries this retrieval ability (\Cref{fig:hotpot_dist:dropdocs}): with documents hidden from the backbone, CPLMs keep 90--97\% of their accuracy with 5 distractors and degrade gracefully as distractors grow, still reaching 34--38\% EM with 500 distractors, on par with or above every model that sees the documents in KV cache, while using a 50$\times$ smaller cache than a Full model. Since documents are encoded independently, their copy keys can also be computed once and reused across queries. In comparison, the last-layer control, with twice the memory of the CP head cache, stays at the level of a model that never sees the documents (14\% EM).

\section{Analysis}\label{sec:analysis}

\subsection{CPLMs Mitigate the LM Head Bottleneck}\label{sec:analysis:softmax}

% \ya{this sub-section is not written to be understood. I tried to parse it, but I am giving up}

\citet{grad_bottleneck} show that the LM head unavoidably compresses the training signal when backpropagating it from $V$-dimensional vocabulary space to $D$-dimensional hidden state space. They further show that the LM head loses as much signal as a random matrix of the same shape: the compression level is essentially dictated by the ratio between $D$ and $V$. As a result, small models fail to learn simple patterns once the vocabulary is large compared to their hidden dimension, even when they should be expressive enough to model them.

The CP head does not escape this random-matrix behavior, but its backward operator has a much more favorable shape. At a given sequence position $t$, the CP head query $W_Q h_t$ goes through a dot-product with the CP head keys $W_K h_{1:t-1}$, before a softmax normalization. This is analogous to the LM head forward, where the CP head keys replace the output embeddings. However, the CP head logits are defined over the $t$ positions of the context rather than over the $V$ tokens of the vocabulary, so the CP head keys have shape $t \times D$ instead of $V \times D$. Since $t \ll V$ for typical contexts, the CP head logit gradient loses a much smaller fraction of its norm than the LM head projection does, and this fraction depends on the sequence position rather than on the vocabulary size.

\begin{figure*}[t]
  \centering
  % ---- shared legend line: both legends on one row, independent of the panels ----
  \includegraphics[height=1.5em]{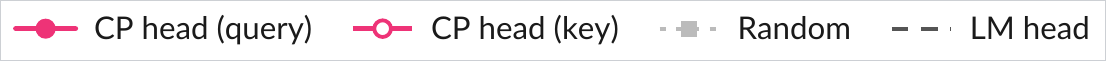}
  \includegraphics[height=1.5em]{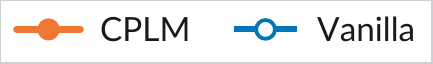}\\
  % ---- three equal-width (1/3) panels ----
  \includegraphics[width=0.3\linewidth]{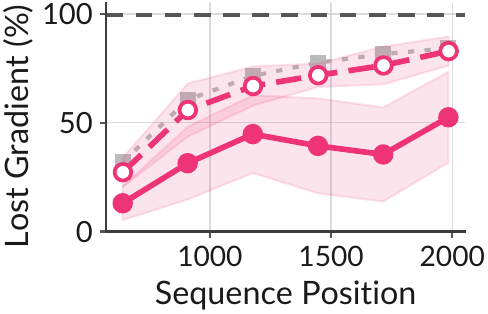}\hfill
  \includegraphics[width=0.3\linewidth]{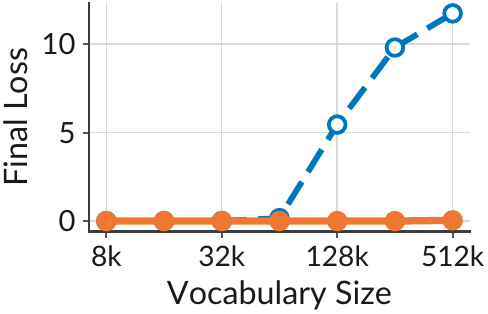}\hfill
  \includegraphics[width=0.3\linewidth]{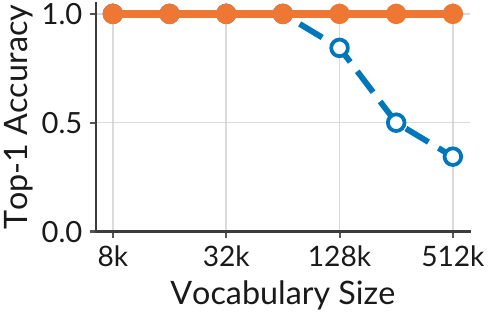}
  \caption{The CP head mitigates the LM head gradient bottleneck \citep{grad_bottleneck}. \textbf{Left:} Fraction of the logit gradient norm that projects to the nullspace of $W_{\{LM, Q,K\}}$ for the 135M full-attention CPLM (\Cref{sec:results:pt}). \emph{Random}: same measurement for a random matrix with the shape of the CP head output matrix. Both CP head pathways let through much more of the signal than the LM head, especially at earlier positions. \textbf{Center and Right:} Accuracy and final validation loss for a 135M model on the SpamLang synthetic task, where CPLMs converge for all vocabulary sizes.}
  \label{fig:bottleneck_analysis}
\end{figure*}

We verify this using the experimental design of \citet{grad_bottleneck}, applied to CPLMs.  \Cref{fig:bottleneck_analysis} (left) shows the fraction of the gradient norm lost in the LM head and in the CP head of our 135M Full vanilla model and CPLM, on FineWeb-Edu sequences of 2{,}048 tokens, and compare it to a random matrix of the same shape as the CP head output matrix. 
The LM heads lose over 99\% of the gradient norm, in line with \citet{grad_bottleneck}. Even if the compression level on the CP head keys pathway remains close to the random reference, their smaller shape lets them conserve up to 85\% of the gradient norm at early positions. This opens a much wider gradient channel than the LM head, especially at early positions.

In \Cref{fig:bottleneck_analysis} (center and right), we additionally train 135M models using both architectures on the SpamLang synthetic task, where each sequence repeats a single uniformly drawn symbol, with a fixed hidden dimension (576) and vocabularies ranging from 8k to 512k tokens. As reported in \citet{grad_bottleneck}, vanilla models fail to reach strong performance for larger vocabularies (34\% accuracy at 512k). CPLMs solve the task at every vocabulary size, showing that the CP head mitigates the LM head bottleneck in this setup.

Further experiments validate that the benefits of CPLMs are tied to the LM head bottleneck (\Cref{app:mechanism}). We conduct a stop-gradient ablation, where we block gradients coming from the CP head to the backbone. This removes most of the benefits of the CP head, showing that backpropagating through the CP head pathway is crucial to the gains of CPLMs. We also pretrain byte-level language models, for which the softmax bottleneck is not active ($D > V=257$); we find that byte-level CPLMs do not improve on vanilla counterparts, hinting that the CP head is only helpful for bottlenecked setups. 

% \subsection{CPLMs Need Fewer Induction Heads}
% \label{sec:analysis:induction}
\subsection{CPLM Backbones Learn Fewer Induction Heads}
\label{sec:analysis:induction}

\begin{figure*}[t]
  \centering
  % --- shared legend, centered on top ---
  \includegraphics[height=1.5em]{imgs/val_nll_curves_legend.pdf}\\
  % --- three panels in one row ---
  \begin{subfigure}[t]{0.35\linewidth}
    \includegraphics[width=\linewidth]{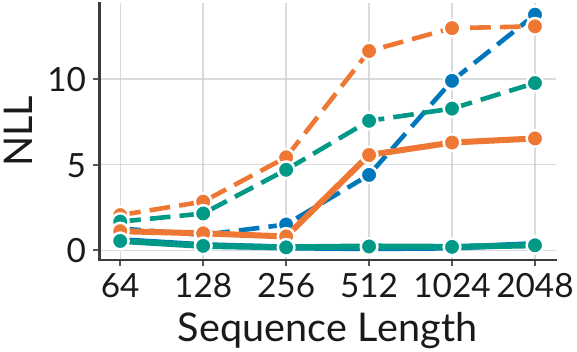}
    \caption{Induction performance}\label{fig:induc_analysis_perf}
  \end{subfigure}\hfill
  \begin{subfigure}[t]{0.35\linewidth}
    \includegraphics[width=\linewidth]{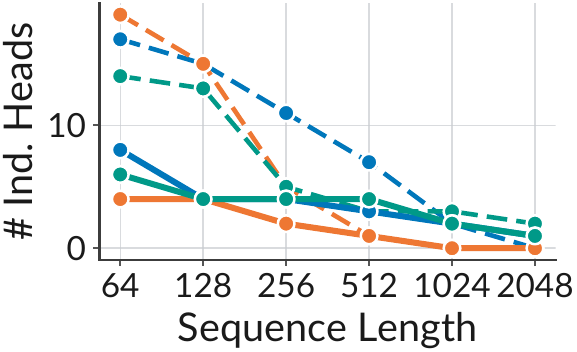}
    \caption{Induction head count}\label{fig:induc_hc}
  \end{subfigure}\hfill
  \begin{subfigure}[t]{0.29\linewidth}
    \includegraphics[width=\linewidth]{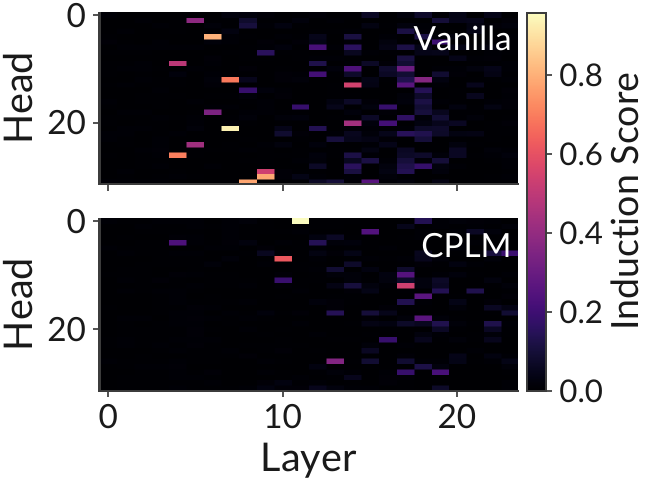}
    \caption{Induction head score}\label{fig:induc_hs}
  \end{subfigure}
  \caption{Induction analysis for 1.7B models. \textbf{Left}: We measure the performance of models when copying a randomly generated sequence; \textbf{Middle}: Total number of triggered induction heads across sequence lengths; \textbf{Right}: Heatmap of induction scores across all heads. Overall, CPLMs have stronger copy abilities while exhibiting far fewer induction heads.}
  \label{fig:induc_analysis}
\end{figure*}

% \ya{this needs some context for induction head and the probe. Olah is not the only one who may read this. If you can't find room for it. Move it to the appendix, and refer to additional analyses  with the gist of each}

% \ya{again: too dense}

% \runin{Induction Heads}
Induction heads~\citep{olsson2022context} are attention heads that implement the
simplest form of in-context copying: if the current token $a$ appeared earlier in the context followed by $b$, an induction head attends to that earlier $b$ and promotes it as the next token. They are a central copying mechanism of Transformer-based language models. 

As the CP head performs this operation explicitly, we ask whether CPLM backbones still learn them. We use the probe of \citet{olsson2022context}: a sequence of $L$ uniformly random tokens, repeated once ($T{=}2L$). The first half is unpredictable, but every token of the second half can be predicted by copying from the first. We measure the NLL on the second half, and the \emph{induction score} of each backbone attention head: its average attention from a token of the second half to the token that followed its first occurrence. Heads scoring above $0.3$ count as induction heads.

% \runin{Results}
CPLMs copy better with far fewer induction heads (\Cref{fig:induc_analysis}).
Full and Hybrid CPLMs predict the repeated half almost perfectly at every length,
while vanilla models degrade as the repeat distance grows. At $T{=}2048$, the Full
vanilla model is even worse than a uniform guess (NLL $13.8$ vs.\ $\ln|V|{=}10.8$),
although the repeat lies within its training context. Meanwhile, CPLM backbones
contain 4 induction heads at $T{=}128$, against 13--15 for vanilla models, and
usually concentrate induction in a single sharp head. The CP head thus takes over
the copying work that vanilla models spread across many heads.

\section{Related Work}\label{sec:related}

\runin{Softmax Bottleneck and Mitigations}
The low rank of the LM head limits the distributions an LM can express~\citep{softmax_bottleneck}. Usual remedies mix several LM heads~\citep{softmax_bottleneck,mixtape,takase2018direct}, or use element-wise non-linearities~\citep{sigsoftmax,pmlr-v97-ganea19a}.
How much this limitation matters in practice is debated~\citep{grivas-etal-2022-low,basri2026the,finlayson2024closing,godey2024why}. These methods target \emph{forward} expressivity and keep an output matrix shared across contexts, and therefore do not address the \emph{backward} bottleneck of \citet{grad_bottleneck}, which the CP head mitigates with a second gradient pathway (\Cref{sec:analysis:softmax}).

\runin{Pointer and Copy Mechanisms}
The CP head is a form of pointer network~\citep{pointer_vinyals}. Pointers have been combined with a vocabulary softmax through a supervised switch~\citep{gulcehre-etal-2016-pointing}, a sigmoid gate~\citep{see-etal-2017-get}, a shared normalizer~\citep{gu-etal-2016-incorporating}, or a sentinel~\citep{w103}, mostly in encoder-decoder or recurrent models and to handle rare words. This line of work largely fell out of favor, as attention-based methods and most notably Transformers~\citep{Vaswani+2017} became dominant. In decoder-only Transformers, where self-attention can already retrieve any previous token, pointers are considered redundant. 

In line with our work, \citet{chang-etal-2023-revisiting-architectures} revisit pointer networks as a remedy to the softmax
bottleneck from the expressivity viewpoint. They partition the output softmax so that in-context tokens are scored by
separate projections of the final hidden state, and continue training GPT-2,
T5 and BART models with these heads. However, the LM head
also bottlenecks optimization~\citep{grad_bottleneck}: it compresses the signal that
shapes the backbone during pretraining, which a head added after pretraining cannot fully
undo. Also, the CPR mechanism presented in~\citet{chang-etal-2023-revisiting-architectures} does not use gating, and instead enforces copy when possible. In this setting, our \texttt{<copy>}-\texttt{<sink>} gating design outperforms their forced-copy approach (\Cref{sec:output-ablations}).

\runin{Externalized Memory}
The CP head is also a context memory. The neural cache~\citep{grave2017improving} and kNN-LM~\citep{khandelwal2020generalization} retrieve from past hidden states at inference, using them as a form of external memory. We compare a tuned neural cache to CPLMs in \Cref{app:ablations:cache}, and show that neural caches bring no gain on our models. TRIME~\citep{zhong-etal-2022-training} trains such a cache jointly with the LM, mainly to exploit long-range and external memories at test time. Its local-memory variant is the closest prior model to CPLMs, but matches past contexts rather than pointing at past tokens, without a gate or learned query and key projections. \citet{zhong-etal-2022-training} also observe that a local cache helps Transformers only when used during training, which the gradient-bottleneck view may explain (\Cref{sec:analysis:softmax}). Memorizing Transformers~\citep{wu2022memorizing} attend to cached per-layer keys and values, RETRO~\citep{borgeaud2022retro} cross-attends to retrieved chunks, CoG~\citep{lan2023copy} copies phrases from a corpus, and several RAG methods precompute per-document KV caches~\citep{lu2025turborag,ma2025blockattention,yao2025cacheblend}. Like the cache, the CP head stores one vector per token, but in a dedicated learned key space. On HotpotQA, with documents hidden from the backbone, these keys alone keep 90--97\% of CPLM accuracy with a $50\times$ smaller total cache than a full KV cache, whereas last-layer states with twice that memory fail (\Cref{sec:res:hotpot}). The CP head also offsets the copying and recall weaknesses of sub-quadratic backbones~\citep{jelassi2024repeat,arora2024zoology}: sliding-window CPLMs match full-attention vanilla LMs.

\section{Conclusion}\label{sec:conclusion}

We introduce Copy-Paste Language Models, which complement the LM head with a lightweight pointer over past positions. Although self-attention can already retrieve any previous token, the copied token must still be decoded through the low-rank LM head, which affects copying abilities. Our CP head adds a second, less compressed prediction and gradient pathway to circumvent the LM head bottleneck. At negligible cost, CPLMs train faster and improve perplexity, downstream performance, supervised fine-tuning and long-context results on every backbone we tested, notably enabling sliding-window models to match full attention with a fraction of the memory.
The CP head only relaxes the LM head bottleneck for tokens that appear in the context, and our experiments stop at 1.7B parameters. Extending the copy-paste mechanism to spans or to external corpora and studying how its benefits evolve at larger scales are natural next steps.

% -----------------------------------------------------------------------
% ICLR statements (do not count toward the page limit)
% -----------------------------------------------------------------------

\subsection*{Acknowledgements}

We thank the members of the Cornell LIL Lab for their helpful comments. This work is partially supported by the National Science Foundation (NSF) under awards OAC-2311521 and DMR-2433348, and NASA under award No. 20-OSTFL20-0053. This work was supported as part of the ``Swiss AI initiative'' by a grant from the Swiss National Supercomputing Centre (CSCS) under project ID a0137 on Alps. NG is supported by an Empire AI Postdoctoral Fellowship. We gratefully acknowledge use of the research computing resources of the Empire AI Consortium, Inc., with support from the State of New York, the Simons Foundation, and the Secunda Family Foundation~\citep{10.1145/3708035.3736070}.

\subsection*{AI use statement}
In this work, we used generative AI tools for literature search, for drafting and editing parts of the paper, including the related work section, proofreading and formatting, and for implementing standard methods such as evaluation and plotting scripts. We have not used generative AI tools to generate research ideas, design experiments, or produce experimental results: all reported numbers come from our own training and evaluation runs. We have reviewed all AI-assisted work: AI-generated code was verified and tested by the authors, AI-drafted or AI-edited text was checked and revised by the authors against our experimental results, and all references were verified manually. We take responsibility for the final content of this work, including text, claims or artifacts produced with the aid of generative AI.

\subsection*{Ethics statement}
This work does not involve human subjects or personal data collection. All models are pretrained on FineWeb-Edu~\citep{fwedu}, a publicly available and filtered web corpus, fine-tuned on the publicly available Tulu-3 data~\citep{lambert2025tulu} and HotpotQA~\citep{yang-etal-2018-hotpotqa}, and evaluated on public benchmarks. As with any language model trained on web data, our models may reproduce biases or factual errors present in their training data. They are small research models that have not undergone safety alignment, and should not be deployed in user-facing applications without further evaluation. Because CPLMs copy from their context more easily, they may also reproduce sensitive or copyrighted content present in their inputs more faithfully. Conversely, sliding-window CPLMs reach the performance of full-attention models with a much smaller memory footprint, which can lower the cost of inference.

% \subsection*{Reproducibility statement}
% The CP head, its gating and its sink mechanism are fully specified in \Cref{sec:meth:cp_head}, and the two-stage decoding procedure in \Cref{sec:output-ablations}. Architectures, data, optimizer and schedules for all pretraining runs are given in \Cref{app:hyperparams}, and multi-seed results at the 135M scale in \Cref{app:seeds}. The output-mechanism baselines we compare to are described in \Cref{sec:output-ablations}. The supervised fine-tuning evaluations are detailed in \Cref{app:sft_evals}, the needle-in-a-haystack protocol in \Cref{app:niah}, and the long-context benchmarks in \Cref{app:long_ctx}. The HotpotQA fine-tuning recipe and the two document-access regimes are described in \Cref{sec:res:hotpot}. All datasets used in this work are publicly available, and we will release our code and trained checkpoints upon publication.


\bibliography{iclr2027_conference}
\bibliographystyle{iclr2027_conference}

\appendix
\crefalias{section}{appendix}\crefalias{subsection}{appendix}\crefalias{subsubsection}{appendix}
\clearpage
\section{Pretraining Hyperparameters}\label{app:hyperparams}

All pretraining runs follow the SmolLM2 recipe~\citep{allal2025smollm2smolgoesbig}, and vanilla models and CPLMs share the exact same hyperparameters, listed in \Cref{tab:pretrain-hparams}. The CP queries and keys use the hidden dimension $D$ unless stated otherwise.

\begin{table*}[ht]
\centering

\begin{tabular}{l ccc}
  \toprule
                               & \textbf{135M} & \textbf{360M} & \textbf{1.7B} \\
  \midrule
  \multicolumn{4}{l}{\emph{Architecture}} \\
  Layers                       & 30       & 32       & 24 \\
  Hidden size                  & 576      & 960      & 2{,}048 \\
  FFN size                     & 1{,}536  & 2{,}560  & 8{,}192 \\
  Attention heads              & 9        & 15       & 32 \\
  KV heads (GQA)               & 3        & 5        & 32 \\
  Head dim                     & 64       & 64       & 64 \\
  Vocabulary size              & 49{,}152 & 49{,}152 & 49{,}152 \\
  Tied embeddings              & Yes      & Yes      & Yes \\
  \midrule
  \multicolumn{4}{l}{\emph{Optimization \& schedule}} \\
  Sequence length              & 2{,}048  & 2{,}048  & 4{,}096 \\
  Batch size (sequences)       & 512      & 512      & 512 \\
  Training steps               & 19{,}000 & 19{,}000 & 95{,}000 \\
  Training tokens              & 20\,B    & 20\,B    & 200\,B \\
  Optimizer                    & AdamW    & AdamW    & AdamW \\
  Peak learning rate           & $3\times10^{-3}$ & $3\times10^{-3}$ & $5\times10^{-4}$ \\
  LR schedule                  & WSD      & WSD      & WSD \\
  Warmup steps                 & 500      & 500      & 2{,}000 \\
  Cooldown steps               & 1{,}900  & 1{,}900  & 4{,}750 \\
  Weight decay                 & 0.01     & 0.01     & 0.01 \\
  Adam $(\beta_1,\beta_2)$     & (0.9, 0.95) & (0.9, 0.95) & (0.9, 0.95) \\
  Gradient clipping            & 1.0      & 1.0      & 1.0 \\
  Precision                    & \texttt{bf16-mixed} & \texttt{bf16-mixed} & \texttt{bf16-mixed} \\
  \bottomrule
\end{tabular}
\caption{Pretraining hyperparameters for the 135M, 360M and 1.7B runs.}
\label{tab:pretrain-hparams}
\end{table*}

Full-attention and hybrid models use YaRN~\citep{peng2024yarn} with a $\times$4 scaling factor from their 4,096-token pretraining context, applied to all full-attention layers: at inference only for the base models, and already during fine-tuning for the SFT models. SW-256 models keep plain RoPE, since they never see positions more than 256 tokens back.

\section{Latency and Memory}\label{app:efficiency}

The CP head adds two $D \times D$ projections and a sink key to the backbone, i.e. 0.5\% more parameters at 1.7B and 9B, and 0.2\% at 30B. The 9B and 30B models are untrained architectures, only used to measure time and memory. We measure what this costs in time and memory on a single NVIDIA B300 GPU in \texttt{bfloat16}, comparing each CPLM with a vanilla model with the same backbone. We focus on a full-attention model for this analysis.

\paragraph{Training.} \Cref{tab:latency_train} reports training throughput and peak memory with sequences of 4{,}096 tokens. The CP head distribution and its contribution to the loss are computed with a fused kernel, so that CPLMs never materialize the $T \times T$ copy distribution over the vocabulary. At 1.7B, CPLMs train at 97\% of the vanilla throughput, and the gap closes with scale: at 9B and 30B, the difference is within 1\%, i.e. at the level of run-to-run variation. This is expected, as the cost of the CP head grows with the hidden dimension and the sequence length but not with depth, so that it becomes a smaller share of each training step as models grow. Peak memory increases by 2.4--3.0\,GB at 9B and 30B (1.7--2.6\%). At 1.7B, computing this contribution over the 32k tokens of the batch at once adds 12.1\,GB, and computing it in chunks of 8{,}192 positions reduces this overhead to 3.4\,GB (+2.7\%).

\begin{table}[ht]
  \centering
  \small
  \begin{tabular*}{\linewidth}{@{\extracolsep{\fill}}lrrrrrr@{}}
  \toprule
   & \multicolumn{3}{c}{Throughput (tokens/s)} & \multicolumn{3}{c}{Peak memory (GB)} \\
  \cmidrule(lr){2-4} \cmidrule(l){5-7}
  Model & Vanilla & CPLM & Ratio & Vanilla & CPLM & Ratio \\
  \midrule
  1.7B & 55{,}862 & 54{,}113 & $\times$0.969 & 125.4 & 137.5 & $\times$1.096 \\
  9B   & 20{,}429 & 20{,}533 & $\times$1.005 & 114.0 & 117.0 & $\times$1.026 \\
  30B  & 5{,}391  & 5{,}342  & $\times$0.991 & 139.9 & 142.3 & $\times$1.017 \\
  \bottomrule
  \end{tabular*}
  \caption{Training cost on one B300 GPU with sequences of 4{,}096 tokens. Ratios are CPLM values divided by vanilla values. At 1.7B, a training step includes the AdamW update (batch of 8 sequences); at 9B and 30B, only the forward and backward passes fit on one GPU.}
  \label{tab:latency_train}
\end{table}

\paragraph{Inference.} At inference time, CPLMs cache one copy key per token, next to the key--value cache. During prefill, the copy keys of the prompt are written to this cache and the copy distribution is only computed for the last position, so that prefill never forms the $T \times T$ copy scores. Each decoding step then adds a single dot product between the copy query and each cached copy key, followed by the mixture with the LM head distribution. We measure generation with a static cache, FlexAttention and a decoding step compiled into a CUDA graph, prefilling prompts of 4k or 16k tokens and greedily generating 256 tokens (\Cref{tab:latency_inference}).

Prefill time is unchanged: all differences are within $\pm 1.3\%$ and go in both directions. A decoding step is 1.0--3.8\% slower. Since decoding is limited by memory bandwidth, the extra time is essentially that of reading the cached copy keys. It is largest when the key--value cache dominates the decoding step (large batches of long sequences), and smallest when reading the weights does (small batches). As prefill comes at no cost, the overhead per request is even smaller: for a prompt of 4k to 16k tokens followed by 256 generated tokens, CPLMs add 0.8--1.7\% to the total time.

The copy cache holds $D$ values per token, against $2\, n_{\text{layers}}\, n_{\text{kv}}\, d_{\text{head}}$ for the key--value cache, i.e. 2.1\% of the key--value cache at 1.7B and 6.2\% at 9B, and decoding memory grows by exactly this amount.

\begin{table}[ht]
  \centering
  \small
  \begin{tabular*}{\linewidth}{@{\extracolsep{\fill}}llrrrrrr@{}}
  \toprule
   & & & \multicolumn{3}{c}{Time} & \multicolumn{2}{c}{Memory (GB)} \\
  \cmidrule(lr){4-6} \cmidrule(l){7-8}
  Model & $T$ & $B$ & Prefill & Decoding step & Request & KV cache & Copy cache \\
  \midrule
  1.7B & 4k  & 1   & $\times$0.993 & $\times$1.018 & $\times$1.017 & 0.80 & 0.02 \\
       & 4k  & 128 & $\times$0.999 & $\times$1.017 & $\times$1.010 & 102.1 & 2.13 \\
       & 16k & 32  & $\times$0.996 & $\times$1.018 & $\times$1.008 & 97.5 & 2.03 \\
  \midrule
  9B   & 4k  & 1   & $\times$0.987 & $\times$1.010 & $\times$1.008 & 0.53 & 0.03 \\
       & 4k  & 128 & $\times$1.000 & $\times$1.038 & $\times$1.014 & 68.1 & 4.25 \\
       & 16k & 32  & $\times$1.001 & $\times$1.027 & $\times$1.011 & 65.0 & 4.06 \\
  \bottomrule
  \end{tabular*}
  \caption{Inference cost on one B300 GPU, for prompts of $T$ tokens and batches of $B$ sequences. Time columns give the ratio of the CPLM time to the vanilla time for prefill, for a decoding step, and for a whole request (prefill and 256 generated tokens), so that values above 1 mean that the CPLM is slower. Memory columns give the size of the key--value cache and of the additional copy cache of the CPLM in \texttt{bfloat16}.}
  \label{tab:latency_inference}
\end{table}

Overall, the cost of CPLMs is of the same small order in training and at inference: at most a few percent of throughput, less than 1\% at 9B and 30B in training, and memory overheads of a few percent.

\section{Ablations}\label{app:ablations}

\subsection{Multi-Seed Robustness}\label{app:seeds}

We repeat the 135M/20B runs with three seeds (42, 43, 44) to confirm the copy
gains exceed seed variance. Each arm uses the same recipe as described in \Cref{app:hyperparams} with a window size of 256 where it applies; the only change
across runs is the random seed. \Cref{tab:seeds} reports final validation
perplexity (mean $\pm$ standard deviation over the three seeds) for each backbone,
with and without the CP head. CPLM lowers perplexity for every backbone,
and each gain is roughly $15$--$40\times$ the seed standard deviation. A one-sided
paired $t$-test gives a p-value $p<10^{-3}$ in all three
cases.

\begin{table}[ht]
\centering
\begin{tabular}{lcccc}
\toprule
Backbone & Vanilla & CPLM & $\Delta$ & $p$ \\
\midrule
Full        & $16.33 \pm 0.03$ & $15.83 \pm 0.02$ & $-0.50$ & $3.8\times10^{-4}$ \\
SW      & $16.89 \pm 0.02$ & $16.02 \pm 0.02$ & $-0.87$ & $9.5\times10^{-5}$ \\
Hybrid 3:1  & $16.37 \pm 0.02$ & $15.84 \pm 0.04$ & $-0.52$ & $2.5\times10^{-4}$ \\
\bottomrule
\end{tabular}
\caption{Final validation perplexity at 135M (mean $\pm$ std over seeds
42/43/44). $\Delta$ is CPLM minus Vanilla.}
\label{tab:seeds}
\end{table}

\subsection{Copy Dimension}\label{app:ablations:copy_dim}

\begin{figure}[ht]
  \centering
  \includegraphics[width=0.45\linewidth]{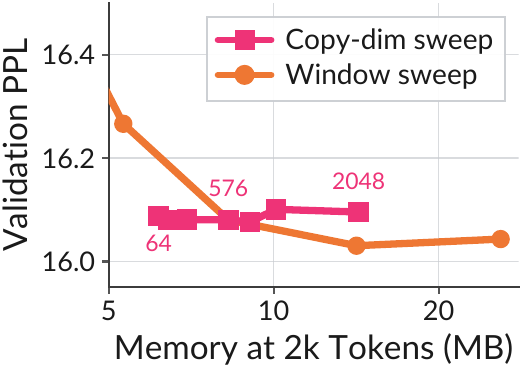}
  \caption{Validation perplexity vs. cache memory at 2k tokens for our 135M sliding-window CPLM ($w=256$), when varying the dimension of the CP queries and keys from 64 to 2{,}048 (\emph{Copy-dim sweep}, default $D=576$), or the backbone window size (\emph{Window sweep}).}
  \label{fig:copydim}
\end{figure}

By default, the CP queries and keys use the hidden dimension $D$. \Cref{fig:copydim} shows that this dimension barely matters: with copy dimensions ranging from 64 to 2{,}048, the validation perplexity of our 135M sliding-window CPLM stays within 0.02 points, while the memory needed to cache the copy keys varies substantially. A copy dimension of 64 thus stores $9\times$ fewer values per token than the default $D=576$ for the same performance. This makes the copy dimension a much cheaper knob than the window size: reducing the window to reach the same memory budget increases perplexity by up to 0.2 points. Small copy dimensions would also further reduce the memory of the copy cache in retrieval settings (\Cref{sec:res:hotpot}). For the sake of simplicity, we use the default $D$ for our larger pretraining runs, and leave the exploration of this phenomenon for future work.

\subsection{360M Scale}\label{app:ablations:360m}

\begin{figure}[ht]
  \centering
  \includegraphics[height=1.5em]{imgs/val_nll_curves_legend.pdf}\\[2pt]
  \includegraphics[width=0.5\linewidth]{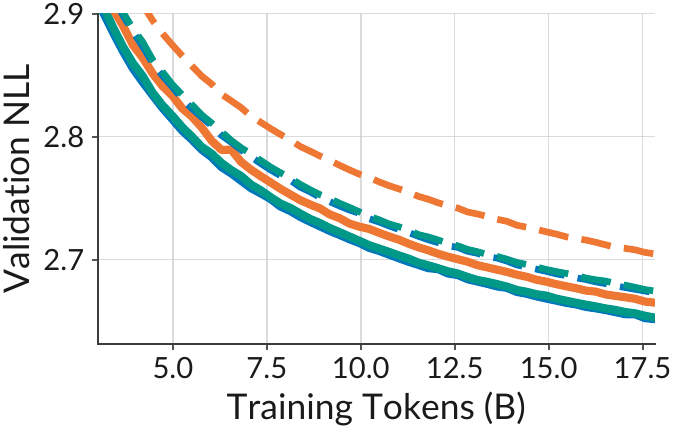}
  \caption{FineWeb-Edu validation loss vs. training tokens (before LR decay) at 360M scale, for $w=256$. Full and Hybrid training curves are essentially superposed.}
  \label{fig:val_nll_curves_360M}
\end{figure}

\Cref{fig:val_nll_curves_360M} reports the training dynamics of our 360M models, trained with the same data and token budget as our 135M models (\Cref{tab:pretrain-hparams}). The results are the same as at 135M and 1.7B scale (\Cref{fig:val_nll_curves}): CPLMs improve over their vanilla counterparts for every backbone, and the SW CPLM outperforms the vanilla Full and Hybrid models.

\subsection{Sink and Window Size}\label{app:ablations:sink}

\begin{figure}[ht]
  \centering
  \includegraphics[width=0.5\linewidth]{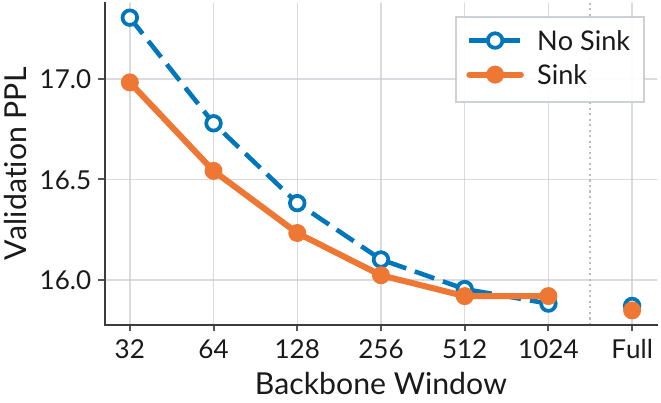}
  \caption{Validation perplexity of 135M CPLMs with and without the sink mechanism, for sliding-window backbones with windows from 32 to 1{,}024 tokens and for a full-attention backbone.}
  \label{fig:sink_window}
\end{figure}

\Cref{fig:sink_window} compares CPLMs with and without the sink across backbone window sizes. The sink consistently helps when the window is small, lowering perplexity by about 0.3 points with a 32-token window and by 0.07 points with a 256-token window. The gap vanishes for windows of 512 tokens or more and for full attention. We hypothesize that, with a small window, the backbone has less information about whether the next token can be found in the context, so that the ability to defer back to the LM head after triggering the CP head becomes more valuable.

\subsection{Gated DeltaNet Pretraining}\label{app:ablations:gdn}

We replace the softmax-attention mixer with a Gated DeltaNet
(GDN) layer \citep{gated_dn}, a linear-attention recurrence whose state is constant in sequence
length, keeping the rest of the backbone, the Copy-Paste head, and the training recipe
identical to the other 135M experiments. Similarly to sliding-window attention, we train two variants---a fully-recurrent GDN model
and a hybrid that keeps one attention layer for every three GDN layers---each with
and without the CP head mechanism, for $20$B tokens on the same doc-masked data.
Even though we match the hyperparameters of the other 135M runs, this comparison is not parameter-matched: the GDN mixer is larger
than an attention block ($1.78$M vs.\ $0.88$M parameters per layer), so the
fully-recurrent model carries about $20\%$ more parameters than the corresponding
attention arm ($161$M vs.\ $134.5$M), which would confound direct comparison.

\begin{figure*}[t]
  \centering
  % --- shared legend, centered on top ---
  \includegraphics[height=1.5em]{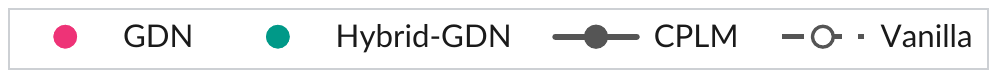}\\
  % --- four panels in one row ---
  \begin{subfigure}[t]{0.35\linewidth}
    \includegraphics[width=\linewidth]{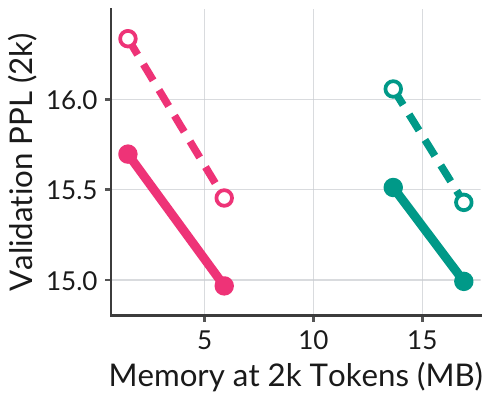}
    \caption{FineWeb-Edu}
  \end{subfigure}
  \begin{subfigure}[t]{0.35\linewidth}
    \includegraphics[width=\linewidth]{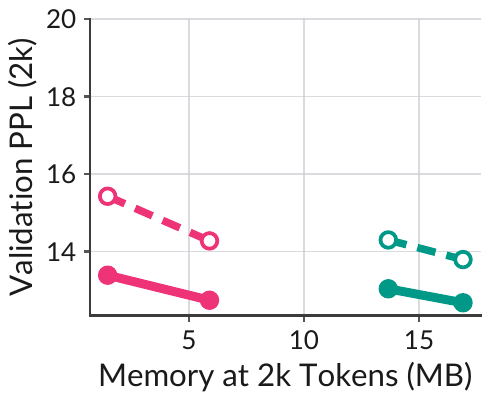}
    \caption{FinePDFs}
  \end{subfigure}
  \begin{subfigure}[t]{0.35\linewidth}
    \includegraphics[width=\linewidth]{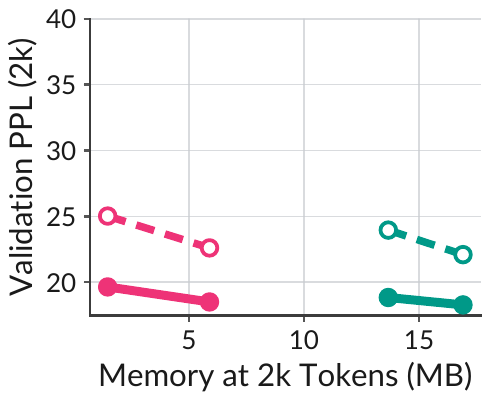}
    \caption{arXiv}
  \end{subfigure}
  \begin{subfigure}[t]{0.35\linewidth}
    \includegraphics[width=\linewidth]{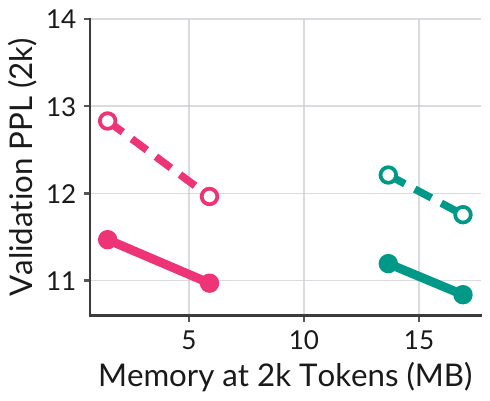}
    \caption{GovReport}
  \end{subfigure}
  \caption{Validation perplexity vs. memory usage at 135M scale for fully-recurrent (GDN) and hybrid (Hybrid-GDN) Gated DeltaNet backbones, with and without the CP head, on in-domain (FineWeb-Edu) and out-of-distribution (FinePDFs, arXiv, GovReport) data. Each architecture is shown with GDN state sizes of 64 and 256, the larger state using more memory.}
  \label{fig:ppl_vs_memory_gdn}
\end{figure*}

\Cref{fig:ppl_vs_memory_gdn} shows that the CP head also benefits recurrent backbones. For both GDN and Hybrid-GDN models, and for both state sizes, CPLMs have a lower perplexity than their vanilla counterparts on every dataset, by about 0.5 points on in-domain FineWeb-Edu and by 1 to 6 points on out-of-distribution data. As with sliding-window attention (\Cref{fig:ppl_vs_memory}), a fully-recurrent GDN CPLM outperforms the vanilla Hybrid-GDN models on out-of-distribution data, while using several times less memory.

\subsection{CP Head Ablations}
\label{sec:output-ablations}

We compare ways of combining a pointer over past tokens with the LM head. Unless stated otherwise, every variant uses the same backbone, and the same queries $q_t$, keys $k_\tau$ and copy attention $\alpha_t(\tau)$ over the context positions $\tau \le t$ as the CP head (\Cref{eq:attn}), with $p_{\mathrm{CP}}(v \mid x_{\le t})$ defined as in \Cref{eq:cp}. Variants only differ in how the LM head and the pointer are combined into the next-token distribution $p(v \mid x_{\le t})$.

\paragraph{Sigmoid gate.} As in pointer-generator networks~\citep{see-etal-2017-get}, the two distributions are mixed with a separate gate $g_t = \sigma(w_g^\top h_t)$:
\begin{equation*}
    p(v \mid x_{\le t}) \ = \ (1 - g_t) \, p_{\mathrm{LM}}(v \mid x_{\le t}) \ + \ g_t \, p_{\mathrm{CP}}(v \mid x_{\le t}),
\end{equation*}
and the \texttt{<copy>} token is not used.

\paragraph{Shared softmax.} Following CopyNet~\citep{gu-etal-2016-incorporating}, the vocabulary logits and the copy scores share a single normalizer:
\begin{gather*}
    p(v \mid x_{\le t}) \ = \ \frac{1}{Z_t} \Big( \exp\big(w_v^\top h_t\big) + \sum_{\tau \le t \,:\, x_\tau = v} \exp\big(k_\tau^\top q_t / \sqrt{D}\big) \Big), \\
    Z_t \ = \ \sum_{u \in \mathcal{V}} \exp\big(w_u^\top h_t\big) + \sum_{\tau \le t} \exp\big(k_\tau^\top q_t / \sqrt{D}\big).
\end{gather*}
There is no explicit gate: the share of copy mass is set by the scale of the copy scores relative to the vocabulary logits, and it grows with the number of visible positions. We keep our dot-product copy scores and omit the selective-read state update of CopyNet.

\paragraph{\citet{chang-etal-2023-revisiting-architectures} partitions.} Following \citet{chang-etal-2023-revisiting-architectures}, the vocabulary is split into partitions, scored with separate projections of the hidden state inside a single softmax. The base logit of token $v$ is $L_V(h_t)^\top w_v$. The context partition (C) rescores the tokens present in the context with $L_C(h_t)^\top w_v$. The pointer partition (P) adds $L_{PD}(h_t)^\top e_v$ for the tokens present in the context, where $e_v$ averages $L_{LD}(h_\tau)$ over the past occurrences $\tau$ of $v$. The reranker (R) rescores the top-20 tokens that are not in the context with $L_R(h_t)^\top w_v$. CP and CPR combine these partitions. All variants are trained with cross-entropy. The original work fine-tunes a pretrained model with a multi-layer augmented hidden state, pointer projections initialized at $10^{-10} I$, and two rerankers. To pretrain from scratch, we use the final hidden state, initialize the pointer projections at $10^{-3} I$ so that the pointer engages within our training budget, and use a single reranker.

\paragraph{Remark on decoding cost.} Our gate is a token of the LM head, so the amount of copying is known before any copy attention is computed. Sampling can therefore proceed in two stages: sample from the LM head over $\mathcal{V} \cup \{\texttt{<copy>}\}$, and only compute the copy attention when \texttt{<copy>} is sampled, falling back to $p_{\mathrm{LM}}$ if the sink is then sampled. This is exact sampling from the mixture of \Cref{eq:proba_sink}. For greedy decoding, copying adds at most $\lambda_t \le c_t$ to the probability of any token, while scaling the LM head distribution by $1 - \lambda_t$. Hence, if the two largest probabilities $p_1 \ge p_2$ of $p_{\mathrm{LM}}(\cdot \mid x_{\le t})$ satisfy $(1 - c_t)(p_1 - p_2) \ge c_t$, the argmax of $p_{\mathrm{LM}}$ is also the argmax of the mixture, and the copy attention can be skipped. The shared softmax and the partitions of \citet{chang-etal-2023-revisiting-architectures} have no such signal: their normalizer or logits depend on the copy scores, so that the copy computation is needed at every decoding step.

\begin{table}[t]
\centering
\small
\setlength{\tabcolsep}{5pt}
\begin{tabular}{llcccc}
\toprule
 & & \multicolumn{2}{c}{Validation} & \multicolumn{2}{c}{FinePDFs @2k} \\
\cmidrule(lr){3-4}\cmidrule(lr){5-6}
Method & Output mechanism & Full & SW-256 & Full & SW-256 \\
\midrule
Vanilla            & LM softmax                                   & 16.34 & 16.88 & 14.25 & 16.52 \\
\midrule
\rowcolor{gray!8}\multicolumn{6}{l}{\textit{\citet{chang-etal-2023-revisiting-architectures}}} \\
\quad C                  & Context partition                            & 16.25 & 16.71 & 13.88 & 15.66 \\
\quad P                  & Pointer partition                            & 16.19 & 16.54 & 13.81 & 14.96 \\
\quad CP                 & Context + pointer                            & 16.19 & 16.53 & 13.64 & 14.91 \\
\quad R                  & Top-$k$ reranker                             & 16.58 & 17.12 & 14.28 & 16.61 \\
\quad CPR                & Context + pointer + reranker                 & 16.33 & 16.62 & 13.75 & 14.85 \\
\rowcolor{gray!8}\multicolumn{6}{l}{\textit{Pointer variants}} \\
\quad Sigmoid gate       & Mixture, separate gate $\sigma(W_g h)$       & 15.94 & 16.20 & 13.39 & 14.25 \\
\quad Shared softmax     & Single normalizer (CopyNet-style)            & 15.88 & 16.09 & 13.32 & \textbf{13.82} \\
\rowcolor{gray!8}\multicolumn{6}{l}{\textit{CPLM}} \\
\quad w/o sink           & \texttt{<copy>}-gated mixture                & 15.87 & 16.10 & 13.32 & 14.13 \\
\quad w/ sink (ours)     & \texttt{<copy>}-gated mixture + sink         & \textbf{15.85} & \textbf{16.02} & \textbf{13.31} & \textbf{13.82} \\
\bottomrule
\end{tabular}
\caption{Output-mechanism ablations at 135M (20B tokens, seed 42). Perplexity
for a full-attention and a sliding-window (256) backbone, on the in-distribution
validation set (final) and on held-out FinePDFs documents (2{,}000 English
documents, first 2{,}048 tokens each, whitespace- and line-break-normalized).
All runs share the backbone, data and training recipe. Across three seeds the
standard deviation of the main arms on validation is 0.015--0.037 PPL
(\Cref{app:seeds}).}
\label{tab:output-ablations}
\end{table}

\paragraph{Results.} \Cref{tab:output-ablations} reports final perplexity.
The copy-gated mixture with a sink matches or slightly improves over the other
pointer variants on both backbones and evaluation sets, and ties with the shared
softmax on FinePDFs with the sliding-window backbone. The partitions of \citet{chang-etal-2023-revisiting-architectures}
remain behind every pointer variant; except for the reranker, which underperforms
the vanilla model, they improve over the vanilla model, with CP the strongest.
The margins between our method and the other pointer variants are small, within
about one to two seed standard deviations (\Cref{app:seeds}). Note that the pointer-generator
gate~\citep{see-etal-2017-get} and CopyNet~\citep{gu-etal-2016-incorporating}
were designed for encoder-decoder models that copy from a source sequence, and we
adapt them here to copying from the past context of a decoder-only LM.
These results are consistent with our framing: our contribution is not the specific
pointer design, but showing that such a head substantially improves LM pretraining
across backbones, mostly by mitigating the LM head gradient bottleneck
(\Cref{sec:analysis:softmax,app:mechanism}). We use the \texttt{<copy>} gate and the sink
for convenience rather than performance. By construction, the gate allows exact two-stage
sampling that only computes the copy attention when needed (see the remark above), and the gate and
sink probabilities expose the decision to copy and the availability of a suitable position,
which we use in our analyses (\Cref{fig:rolling,fig:copy_bins}).

\subsection{Comparison with a Neural Cache}\label{app:ablations:cache}

A natural question is whether the gains of CPLMs could be obtained without training, by adding a context cache to a vanilla LM. We compare against the neural cache of \citet{grave2017improving}, applied to each vanilla backbone. At position $t$, the cache scores every earlier position $\tau$ by the similarity between the final hidden states $h_t$ and $h_\tau$, and proposes the token that followed $\tau$:
\begin{equation*}
    p_{\mathrm{cache}}(v \mid x_{\le t}) \ \propto \sum_{\tau < t \,:\, x_{\tau+1} = v} \exp\big(\theta\, h_t^\top h_\tau\big).
\end{equation*}
The next-token distribution is then $(1-\gamma)\, p_{\mathrm{LM}} + \gamma\, p_{\mathrm{cache}}$. We tune $\theta$ and $\gamma$ for each backbone on 100 held-out FineWeb-Edu documents, and evaluate on 100 different documents of 2{,}048 tokens (\Cref{tab:neural-cache}).

\begin{table}[ht]
  \centering
  \small
  \begin{tabular}{lccc}
    \toprule
    Attention & Vanilla & Vanilla + Neural cache & CPLM \\
    \midrule
    Full   & 9.17 & 9.17 & \textbf{9.06} \\
    Hybrid 3:1 & 9.20 & 9.20 & \textbf{9.06} \\
    SW-256     & 9.65 & 9.63 & \textbf{9.24} \\
    \bottomrule
  \end{tabular}
  \caption{FineWeb-Edu perplexity of our 1.7B models (100 documents of 2{,}048 tokens). The neural cache~\citep{grave2017improving} is added to the vanilla model, with $\theta$ and $\gamma$ tuned on separate FineWeb-Edu documents.}
  \label{tab:neural-cache}
\end{table}

The tuned neural cache barely changes the perplexity of the vanilla models: the best mixture weight is $\gamma \le 0.002$ for all three backbones, i.e. the cache is almost switched off. CPLMs, which learn when and where to copy, lower perplexity by 0.11 (Full), 0.14 (Hybrid) and 0.41 (SW). The gains of CPLMs therefore come from training the copy mechanism jointly with the LM, not from having access to a context cache.

\section{SFT Evaluation}
\label{app:sft_evals}

SFT uses sequences of up to 16k tokens. All SFT evaluations use the chat template seen during fine-tuning, except MATH,
which uses its standard plain prompt. Full and Hybrid models use YaRN RoPE scaling
(factor 4); SW-256 models use standard RoPE.

\paragraph{Perplexity.}
\emph{Tulu-3} is the validation perplexity at the end of SFT.
\emph{Nemotron-3} is the perplexity on assistant tokens at positions 7.5k--8.5k, within the 16k-token context used for SFT, of
4{,}000 randomly sampled conversations from Nemotron-SFT-Math-v4, with reasoning
traces included in the assistant turns.

\paragraph{MATH.}
Level~1--2 problems from four subjects of MATH~\citep{hendrycksmath2021}
(Algebra, Counting \& Probability, Number Theory, Prealgebra; 861 problems), with
the Minerva 4-shot prompt~\citep{lewkowycz2022solving}. We only use levels 1 and 2, as harder problems are out of reach for models of this size. We report pass@4 over 8
samples (temperature 0.7, top-$p$ 0.9).

\paragraph{GSM8K.}
5-shot GSM8K~\citep{cobbe2021gsm8k}, pass@4 over 8 samples (temperature 0.7,
top-$p$ 0.9). The answer is the last number in the completion.

\paragraph{HumanEval and MBPP.}
Zero-shot HumanEval~\citep{chen2021evaluating} and MBPP~\citep{austin2021program}:
the task is given in the user turn and the model writes the full function. We
report pass@10 over 20 samples (temperature 0.8, top-$p$ 0.95).

\paragraph{IFEval.}
Zero-shot IFEval~\citep{zhou2023instructionfollowingevaluationlargelanguage} with a repetition penalty of 1.1 and
nucleus sampling (temperature 0.9, top-$p$ 0.95). We report the mean of the
prompt- and instruction-level accuracies under both strict and loose criteria.

\begin{figure}[t]
  \centering
  \begin{subfigure}[t]{0.49\linewidth}
    \includegraphics[width=\linewidth]{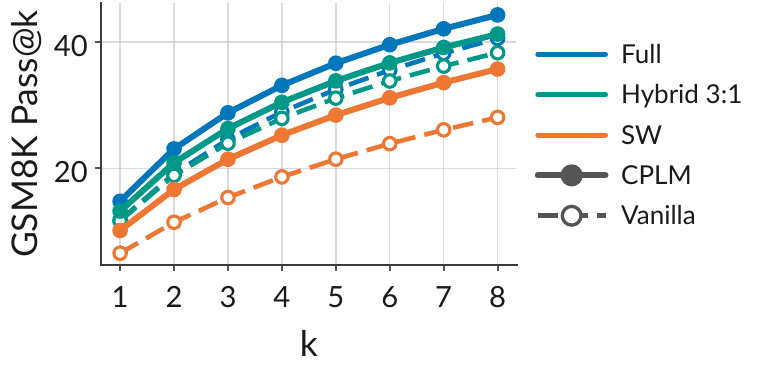}
    \caption{Pass@$k$}\label{fig:gsm8k:passk}
  \end{subfigure}\hfill
  \begin{subfigure}[t]{0.49\linewidth}
    \includegraphics[width=\linewidth]{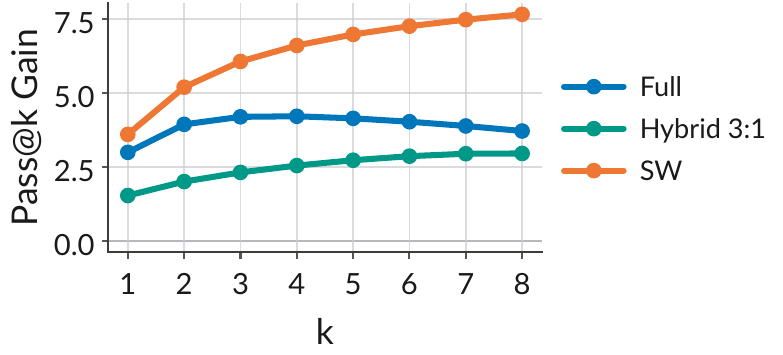}
    \caption{Pass@$k$ gain of CPLMs}\label{fig:gsm8k:delta}
  \end{subfigure}
  \caption{GSM8K pass@$k$ of our 1.7B SFT models for $k$ from 1 to 8, estimated from 8 samples (temperature 0.7, top-$p$ 0.9). \textbf{(a)} Pass@$k$ of vanilla models and CPLMs. \textbf{(b)} Difference between each CPLM and its vanilla counterpart, in points.}
  \label{fig:gsm8k_passk}
\end{figure}

\paragraph{Pass@$k$ consistency.}
We find our pass@$k$ results to be consistent across $k$. For instance, on GSM8K (\Cref{fig:gsm8k_passk}), CPLMs outperform their vanilla counterparts at every $k$ from 1 to 8, including pass@1, where they improve accuracy by 1.6 to 3.6 points. The gain grows with $k$ for the SW and Hybrid CPLMs, reaching 7.7 points at pass@8 for the SW CPLM, while it stays around 4 points for the Full CPLM.

\section{Needle-in-a-Haystack}
\label{app:niah}

\paragraph{Setup.}
We build each prompt from 15 neutral filler sentences, repeated to fill the
context. A needle (``The special magic number is: $N$.'', with $N$ a random
7-digit integer) is inserted at relative depth
$d \in \{0, \tfrac{1}{7}, \dots, 1\}$. The prompt ends with the question
``What is the special magic number?'' We test context lengths
$L \in \{1\text{k}, 2\text{k}, 4\text{k}, 8\text{k}, 16\text{k}, 32\text{k},
64\text{k}\}$ and use 200 examples per $(d, L)$ cell. The model decodes
greedily for up to 16 tokens. An answer counts as correct if the output
contains $N$. For the Full and Hybrid 3:1 models, we apply YaRN ($\times 4$)
to the full-attention layers at every length, so each model uses a single
configuration across the sweep. Sliding-window layers keep the base RoPE,
since they never see a relative offset above 255.

\begin{figure}[t!]
  \centering
  \includegraphics[width=\linewidth]{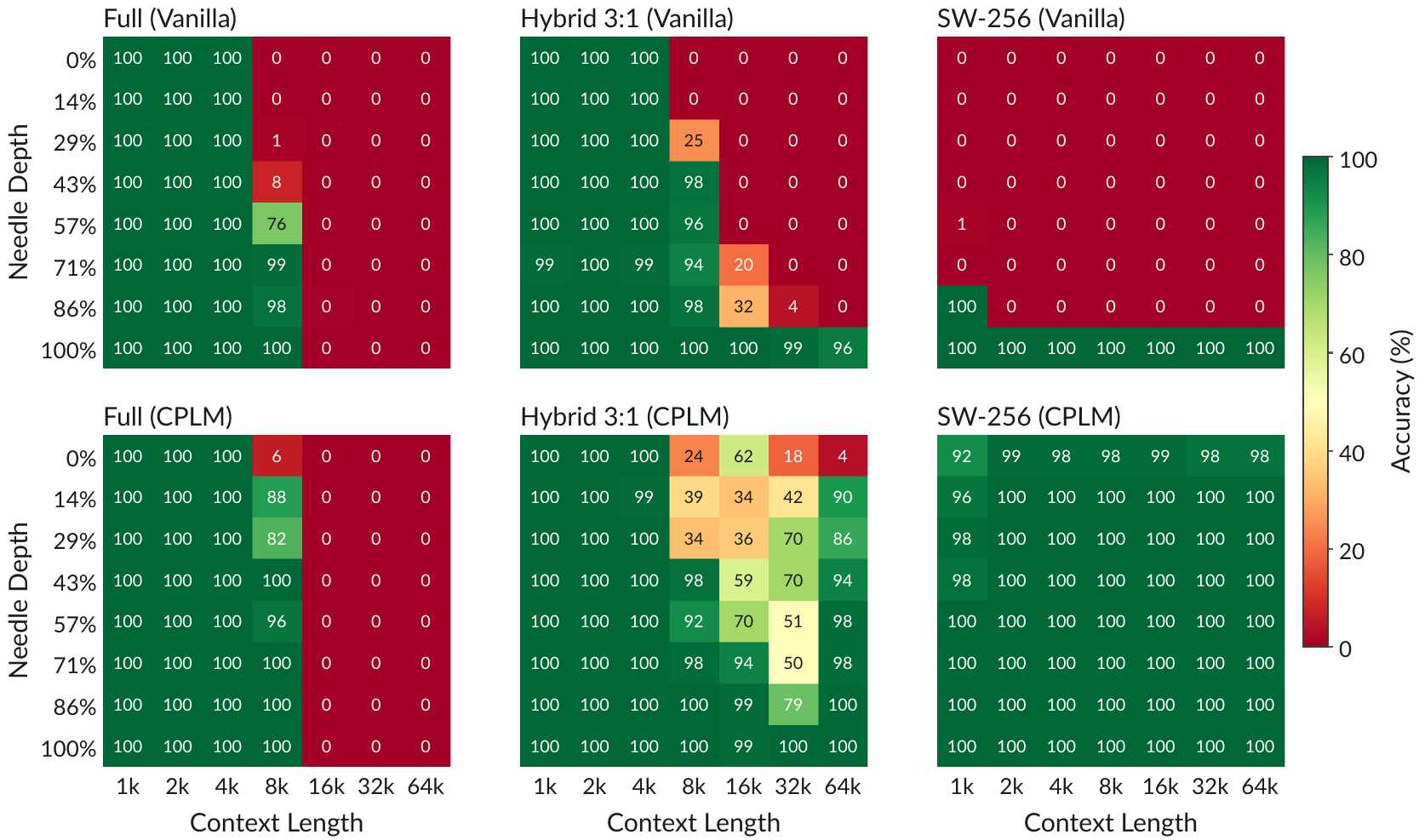}
  \caption{Needle-in-a-haystack accuracy (\%) by needle depth (rows) and
  context length (columns). Top: Vanilla. Bottom: CPLM. Each cell averages 200
  examples. Average over the full grid, Vanilla / CPLM: Full 49.7 / 54.8;
  Hybrid 3:1 58.2 / 83.6; SW-256 14.3 / 99.4.}
  \label{fig:niah}
\end{figure}

\paragraph{Results.}
Up to the 4k pretraining context length, Full and Hybrid models score close to 100\%
at every depth, with or without CPLM. The architectures differ beyond that
point.

\emph{SW-256.} The Vanilla model only finds the needle when it is inside the
last 256 tokens. That is the 100\%-depth row, plus the 86\% row at 1k. Every
other cell is at 0, so accuracy is stuck at $1/8$ of the depths for
$L \geq 2$k. With CPLM, the same backbone gets 98--100\% in almost every cell
up to 64k. The only drop is a small one at depth 0 and at 1k.

\emph{Hybrid 3:1.} Past 8k, the Vanilla Hybrid behaves like a sliding-window
model. It still finds needles at the end of the context, but accuracy on
earlier needles falls to 0. Its average per length drops from 64 at 8k to 12
at 64k. With CPLM, 60--84\% accuracy per length remains from 16k to 64k, and
the deeper needles are recovered best. The shallowest needles
($d \leq 2/7$) are the hardest.

\emph{Full.} Both Full models fail completely from 16k on, even with YaRN.
Retrieval does not generalize much beyond $2\times$ the training length.
CPLM helps in the transition region: at 8k, accuracy rises from 48 to 84, and
Vanilla's weak rows at shallow-to-middle depths are mostly recovered. CPLM
does not move the failure point.

In summary, CPLM helps most when the architecture cannot attend over the
full context. When the backbone's own attention fails beyond the window or
past the training length, the CP head pathway keeps long-range retrieval
working.

\section{Long-Context Evaluation}\label{app:long_ctx}

\Cref{tab:longbench_base} details the LongBench results of our base 1.7B models. The largest gains come from TREC, a few-shot classification task where the answer labels can be copied from the in-context demonstrations: CPLMs improve it by $+16.0$, $+29.0$ and $+24.5$ points for Full, SW and Hybrid models, respectively. However, CPLMs also improve on tasks where copying is not crucial. On the three summarization tasks, whose outputs are abstractive, SW and Hybrid CPLMs improve over their vanilla counterparts on every task, by $+3.6$ to $+6.2$ points, although the Full CPLM is behind its vanilla counterpart on GovReport and QMSum.

\paragraph{LongBench after SFT.} \Cref{tab:longbench_sft} reports the same benchmark after supervised fine-tuning. CPLMs still improve the average score for every backbone ($+0.6$, $+2.7$ and $+1.9$ points for Full, SW and Hybrid models), but the gaps are smaller than for base models and less uniform across tasks. SW and Hybrid CPLMs keep large gains on TREC, while the Full CPLM is behind its vanilla counterpart on this task. Summarization gains are more consistent: CPLMs improve on GovReport and MultiNews for every backbone. Conversely, the SW CPLM loses ground on SAMSum and on code completion after fine-tuning.

\paragraph{RULER and BABILong by context length.} \Cref{tab:longctx_length_full} extends the synthetic evaluations of \Cref{tab:longctx} to 64k tokens. For Full and Hybrid models, CPLMs are better at almost every length up to 16k, and the Hybrid CPLM keeps an advantage on RULER at 32k. Beyond 16k, both Full models fail completely, which shows that the CP head does not compensate for the length generalization failure of the backbone. The SW CPLM improves RULER accuracy at every length, from $10.2$ to $36.1$ at 4k and from $5.0$ to $18.9$ at 64k, while both SW models remain close to the floor on BABILong beyond 2k tokens.

\paragraph{Per-task results at 4k.} \Cref{tab:ruler_tasks_4k,tab:babilong_tasks_4k} break down RULER and BABILong at 4k tokens. For the SW CPLM, the RULER gain mostly comes from single-needle retrieval, which it solves almost perfectly (S1 and S2) while its vanilla counterpart cannot see beyond its window, and from word extraction tasks (CWE and FWE). For Full and Hybrid models, CPLMs improve on single and multi-key retrieval, multi-value retrieval and QA, but not on every task: the Full CPLM is behind on FWE and SQuAD, and the Hybrid CPLM on CWE and FWE. The hardest multi-key tasks (MK2 and MK3) remain near the floor for all models. On BABILong, the Full CPLM improves on all five tasks ($18.6 \rightarrow 32.6$ on average), while the Hybrid CPLM is on par with its vanilla counterpart on average, and both SW models are close to the floor.

\begin{table*}[t]
\centering
\small
\resizebox{\linewidth}{!}{\
\begin{tabular}{lccccccccccccccccc}
\toprule
 & \multicolumn{3}{c}{Single-Doc QA} & \multicolumn{3}{c}{Multi-Doc QA} & \multicolumn{3}{c}{Summarization} & \multicolumn{3}{c}{Few-shot} & \multicolumn{2}{c}{Synthetic} & \multicolumn{2}{c}{Code} &  \\
\cmidrule(lr){2-4}\cmidrule(lr){5-7}\cmidrule(lr){8-10}\cmidrule(lr){11-13}\cmidrule(lr){14-15}\cmidrule(lr){16-17}
 & \rotatebox{80}{NarrativeQA} & \rotatebox{80}{Qasper} & \rotatebox{80}{MultiFieldQA} & \rotatebox{80}{HotpotQA} & \rotatebox{80}{2WikiMQA} & \rotatebox{80}{MuSiQue} & \rotatebox{80}{GovReport} & \rotatebox{80}{QMSum} & \rotatebox{80}{MultiNews} & \rotatebox{80}{TREC} & \rotatebox{80}{TriviaQA} & \rotatebox{80}{SAMSum} & \rotatebox{80}{PassageCount} & \rotatebox{80}{PassageRet} & \rotatebox{80}{LCC} & \rotatebox{80}{RepoBench-P} & \rotatebox{0}{\textbf{Avg.}} \\
\midrule
Full & 1.0 & 3.7 & 12.4 & 6.3 & 9.0 & 3.1 & \textbf{16.6} & \textbf{7.8} & 11.1 & 39.0 & 20.8 & 2.2 & 0.1 & \textbf{3.3} & 15.7 & 15.0 & 10.4 \\
\quad + CP & \textbf{2.6} & \textbf{5.1} & \textbf{12.9} & \textbf{6.4} & \textbf{9.6} & \textbf{4.1} & 11.0 & 6.2 & \textbf{13.2} & \textbf{55.0} & \textbf{26.0} & \textbf{4.0} & \textbf{1.0} & 0.6 & \textbf{20.0} & \textbf{17.1} & \textbf{12.2}\,\gain{+1.7} \\
\midrule
SW-256 & 0.7 & 1.4 & 6.9 & 1.6 & 6.0 & 1.0 & 5.5 & 5.6 & 8.1 & 25.5 & \textbf{24.0} & 0.3 & 0.4 & 2.4 & 14.5 & 15.1 & 7.4 \\
\quad + CP & \textbf{2.7} & \textbf{5.8} & \textbf{14.7} & \textbf{4.0} & \textbf{6.2} & \textbf{3.0} & \textbf{9.1} & \textbf{11.1} & \textbf{12.1} & \textbf{54.5} & 22.0 & \textbf{14.5} & \textbf{0.7} & \textbf{3.4} & \textbf{27.2} & \textbf{22.7} & \textbf{13.4}\,\gain{+5.9} \\
\midrule
Hybrid 3:1 & 0.9 & 2.1 & \textbf{10.8} & \textbf{4.5} & \textbf{8.4} & \textbf{3.1} & 10.5 & 4.7 & 10.0 & 36.0 & 25.2 & 0.2 & 0.5 & 0.9 & 17.2 & 17.7 & 9.5 \\
\quad + CP & \textbf{1.9} & \textbf{5.6} & 8.6 & 3.1 & 5.3 & 1.7 & \textbf{15.9} & \textbf{10.9} & \textbf{15.1} & \textbf{60.5} & \textbf{27.7} & \textbf{17.2} & \textbf{0.9} & \textbf{2.5} & \textbf{24.3} & \textbf{27.0} & \textbf{14.3}\,\gain{+4.7} \\
\bottomrule
\end{tabular}}
\caption{\textbf{LongBench (English, 16 tasks), base models.} All 1.7B models, full-length contexts, no instruction tuning. RoPE: YaRN$\times$4 for Full and Hybrid, base RoPE for SW-256 (YaRN degrades the purely-local arms). \textbf{Bold} marks the better of each vanilla/CPLM pair; \gain{green} is the CP gain in average score.}
\label{tab:longbench_base}
\end{table*}

\begin{table*}[t]
\centering
\small
\resizebox{\linewidth}{!}{\
\begin{tabular}{lccccccccccccccccc}
\toprule
 & \multicolumn{3}{c}{Single-Doc QA} & \multicolumn{3}{c}{Multi-Doc QA} & \multicolumn{3}{c}{Summarization} & \multicolumn{3}{c}{Few-shot} & \multicolumn{2}{c}{Synthetic} & \multicolumn{2}{c}{Code} &  \\
\cmidrule(lr){2-4}\cmidrule(lr){5-7}\cmidrule(lr){8-10}\cmidrule(lr){11-13}\cmidrule(lr){14-15}\cmidrule(lr){16-17}
 & \rotatebox{80}{NarrativeQA} & \rotatebox{80}{Qasper} & \rotatebox{80}{MultiFieldQA} & \rotatebox{80}{HotpotQA} & \rotatebox{80}{2WikiMQA} & \rotatebox{80}{MuSiQue} & \rotatebox{80}{GovReport} & \rotatebox{80}{QMSum} & \rotatebox{80}{MultiNews} & \rotatebox{80}{TREC} & \rotatebox{80}{TriviaQA} & \rotatebox{80}{SAMSum} & \rotatebox{80}{PassageCount} & \rotatebox{80}{PassageRet} & \rotatebox{80}{LCC} & \rotatebox{80}{RepoBench-P} & \rotatebox{0}{\textbf{Avg.}} \\
\midrule
Full & 12.8 & 20.7 & 24.5 & \textbf{27.1} & \textbf{23.6} & 6.5 & 22.2 & \textbf{20.5} & 19.0 & \textbf{15.5} & 66.7 & 27.7 & 0.0 & 4.5 & 19.5 & 21.3 & 20.7 \\
\quad + CP & \textbf{14.8} & \textbf{23.6} & \textbf{27.4} & 23.7 & 18.0 & \textbf{10.7} & \textbf{24.7} & 19.8 & \textbf{22.5} & 8.0 & \textbf{68.1} & \textbf{30.4} & \textbf{1.5} & 4.5 & \textbf{20.9} & \textbf{23.7} & \textbf{21.4}\,\gain{+0.6} \\
\midrule
SW-256 & 10.0 & 9.8 & 16.5 & 14.8 & \textbf{21.2} & \textbf{8.1} & 13.7 & 19.3 & 12.4 & 22.0 & 40.2 & \textbf{17.2} & 0.5 & \textbf{5.5} & \textbf{16.9} & \textbf{18.9} & 15.4 \\
\quad + CP & \textbf{10.8} & \textbf{11.9} & \textbf{20.2} & \textbf{22.5} & 18.0 & 5.7 & \textbf{18.6} & \textbf{20.5} & \textbf{14.7} & \textbf{59.0} & \textbf{46.5} & 6.2 & \textbf{2.4} & 4.2 & 14.5 & 14.3 & \textbf{18.1}\,\gain{+2.7} \\
\midrule
Hybrid 3:1 & \textbf{14.7} & 18.3 & 25.3 & 24.7 & 24.0 & \textbf{10.1} & 22.1 & \textbf{20.2} & 14.8 & 1.0 & \textbf{64.1} & 23.8 & 0.1 & 5.0 & 17.2 & \textbf{22.9} & 19.3 \\
\quad + CP & 13.0 & \textbf{21.5} & \textbf{28.4} & \textbf{24.9} & 24.0 & 8.1 & \textbf{24.8} & 19.0 & \textbf{18.0} & \textbf{20.5} & 60.1 & \textbf{29.1} & \textbf{2.4} & 5.0 & \textbf{22.0} & 17.1 & \textbf{21.1}\,\gain{+1.9} \\
\bottomrule
\end{tabular}}
\caption{\textbf{LongBench (English, 16 tasks), after SFT.} All 1.7B models, instruction-tuned at lr $1{\times}10^{-4}$, 16k contexts. \textbf{Bold} marks the better of each vanilla/CPLM pair; \gain{green} is the CP gain in average score.}
\label{tab:longbench_sft}
\end{table*}

\begin{table*}[t]
\centering
\resizebox{\textwidth}{!}{%
\begin{tabular}{llccccccccccccccc}
\toprule
 & & \multicolumn{6}{c}{RULER} & \multicolumn{9}{c}{BABILong} \\
\cmidrule(lr){3-8} \cmidrule(lr){9-17}
Attention & Model & 4k & 8k & 16k & 32k & 64k & \textbf{Avg} & 0k & 1k & 2k & 4k & 8k & 16k & 32k & 64k & \textbf{Avg} \\
\midrule
Full & Vanilla & 50.1 & 36.3 & 14.5 & 0.0 & 0.0 & 20.2 & \textbf{55.0} & 44.6 & 35.0 & 18.6 & 4.4 & 0.6 & 0.0 & 0.0 & 19.8 \\
 & CPLM & \textbf{54.9} & \textbf{38.6} & \textbf{25.2} & 0.0 & 0.0 & \textbf{23.7} & 52.6 & \textbf{49.6} & \textbf{40.8} & \textbf{32.6} & \textbf{12.6} & \textbf{4.0} & 0.0 & 0.0 & \textbf{24.0} \\
\midrule
Hybrid 3:1 & Vanilla & 50.6 & 32.1 & 16.9 & 7.1 & \textbf{3.6} & 22.1 & 56.4 & 38.8 & 29.0 & 26.8 & 11.4 & 3.6 & 0.6 & 0.2 & 20.8 \\
 & CPLM & \textbf{53.5} & \textbf{40.8} & \textbf{19.2} & \textbf{11.7} & 2.8 & \textbf{25.6} & \textbf{57.6} & \textbf{47.4} & \textbf{42.8} & \textbf{27.4} & \textbf{16.2} & \textbf{5.8} & \textbf{1.4} & 0.2 & \textbf{24.8} \\
\midrule
SW-256 & Vanilla & 10.2 & 7.5 & 5.9 & 5.5 & 5.0 & 6.8 & 46.2 & 18.2 & 10.2 & 5.2 & 1.8 & \textbf{1.0} & 0.4 & 0.6 & 10.5 \\
 & CPLM & \textbf{36.1} & \textbf{27.6} & \textbf{23.9} & \textbf{21.2} & \textbf{18.9} & \textbf{25.5} & \textbf{49.6} & \textbf{22.6} & \textbf{13.0} & \textbf{8.4} & \textbf{2.2} & 0.6 & \textbf{0.8} & 0.6 & \textbf{12.2} \\
\bottomrule
\end{tabular}}
\caption{RULER and BABILong accuracy (\%) by context length for the 1.7B base models. RULER: mean over its 13 tasks; BABILong: mean over qa1--qa5. Avg is the mean over all lengths shown (4k--64k for RULER, 0k--64k for BABILong). Full and Hybrid use YaRN ($\times$4 from 4096); SW-256 uses plain RoPE. \textbf{Bold} marks the better model of each pair.}
\label{tab:longctx_length_full}
\end{table*}

\begin{table*}[t]
\centering
\resizebox{\textwidth}{!}{%
\begin{tabular}{llcccccccccccccc}
\toprule
 & & \multicolumn{8}{c}{Retrieval (NIAH)} & Tracing & \multicolumn{2}{c}{Aggregation} & \multicolumn{2}{c}{QA} & \\
\cmidrule(lr){3-10} \cmidrule(lr){11-11} \cmidrule(lr){12-13} \cmidrule(lr){14-15}
Attention & Model & S1 & S2 & S3 & MK1 & MK2 & MK3 & MQ & MV & VT & CWE & FWE & SQuAD & HotpotQA & \textbf{Avg} \\
\midrule
Full & Vanilla & 99.0 & 100.0 & 68.0 & 62.0 & \textbf{6.0} & \textbf{9.0} & \textbf{83.0} & 64.8 & 0.0 & 34.7 & \textbf{53.3} & \textbf{43.0} & 28.0 & 50.1 \\
 & CPLM & \textbf{100.0} & 100.0 & \textbf{85.0} & \textbf{67.0} & 4.0 & 8.0 & 80.8 & \textbf{72.2} & \textbf{26.6} & \textbf{60.8} & 40.3 & 39.0 & \textbf{30.0} & \textbf{54.9} \\
\midrule
Hybrid 3:1 & Vanilla & \textbf{100.0} & 97.0 & 62.0 & 62.0 & 6.0 & 5.0 & \textbf{82.8} & 73.2 & \textbf{9.0} & \textbf{51.6} & \textbf{42.3} & 38.4 & 29.0 & 50.6 \\
 & CPLM & 99.0 & \textbf{100.0} & \textbf{70.0} & \textbf{90.0} & \textbf{8.0} & \textbf{8.0} & 82.5 & \textbf{79.0} & 7.0 & 38.9 & 38.0 & \textbf{42.3} & \textbf{33.0} & \textbf{53.5} \\
\midrule
SW-256 & Vanilla & 8.0 & 9.0 & 3.0 & 8.0 & 3.0 & 1.0 & 5.2 & 3.2 & \textbf{0.4} & 16.0 & 26.0 & 31.3 & 19.0 & 10.2 \\
 & CPLM & \textbf{100.0} & \textbf{99.0} & \textbf{36.0} & \textbf{39.0} & 3.0 & \textbf{7.0} & \textbf{16.5} & \textbf{20.2} & 0.0 & \textbf{44.4} & \textbf{48.0} & \textbf{32.0} & \textbf{24.0} & \textbf{36.1} \\
\bottomrule
\end{tabular}}
\caption{RULER per-task accuracy (\%) at 4k context for the 1.7B base models. NIAH: single-needle (S1--S3), multi-key (MK1--MK3), multi-query (MQ), multi-value (MV); VT: variable tracking; CWE/FWE: common/frequent word extraction. Avg is the mean over the 13 tasks. Full and Hybrid use YaRN ($\times$4 from 4096); SW-256 uses plain RoPE. \textbf{Bold} marks the better model of each pair.}
\label{tab:ruler_tasks_4k}
\end{table*}

\begin{table}[t]
\centering
\begin{tabular}{llcccccc}
\toprule
Attention & Model & qa1 & qa2 & qa3 & qa4 & qa5 & \textbf{Avg} \\
\midrule
Full & Vanilla & 19.0 & 9.0 & 17.0 & 33.0 & 15.0 & 18.6 \\
 & CPLM & \textbf{31.0} & \textbf{21.0} & \textbf{34.0} & \textbf{34.0} & \textbf{43.0} & \textbf{32.6} \\
\midrule
Hybrid 3:1 & Vanilla & 20.0 & \textbf{12.0} & 23.0 & \textbf{43.0} & 36.0 & 26.8 \\
 & CPLM & \textbf{30.0} & 6.0 & \textbf{29.0} & 30.0 & \textbf{42.0} & \textbf{27.4} \\
\midrule
SW-256 & Vanilla & 4.0 & 0.0 & 5.0 & 5.0 & 12.0 & 5.2 \\
 & CPLM & \textbf{6.0} & \textbf{2.0} & \textbf{10.0} & \textbf{6.0} & \textbf{18.0} & \textbf{8.4} \\
\bottomrule
\end{tabular}
\caption{BABILong per-task accuracy (\%) at 4k context for the 1.7B base models. qa1--qa3: one, two and three supporting facts; qa4: two-argument relations; qa5: three-argument relations. Full and Hybrid use YaRN ($\times$4 from 4096); SW-256 uses plain RoPE. \textbf{Bold} marks the better model of each pair.}
\label{tab:babilong_tasks_4k}
\end{table}

\section{Analyzing CPLMs}\label{app:analysis}

\subsection{Copy Probability and Prediction Quality}\label{app:additional:copy_nll}

\begin{figure*}[ht]
  \centering
  \includegraphics[height=1.5em]{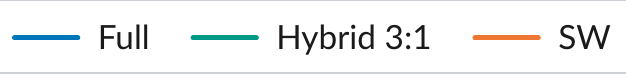}\\[2pt]
  \begin{subfigure}[t]{0.32\linewidth}
    \includegraphics[width=\linewidth]{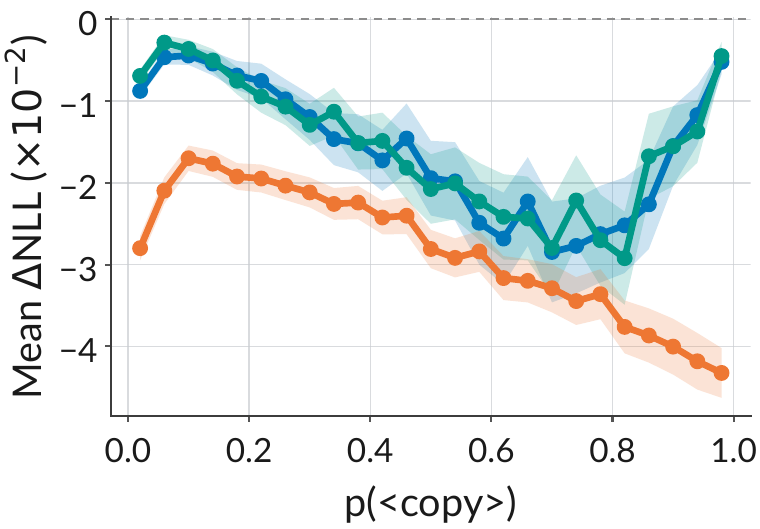}
    \caption{NLL gap}\label{fig:copy_bins:nll}
  \end{subfigure}\hfill
  \begin{subfigure}[t]{0.32\linewidth}
    \includegraphics[width=\linewidth]{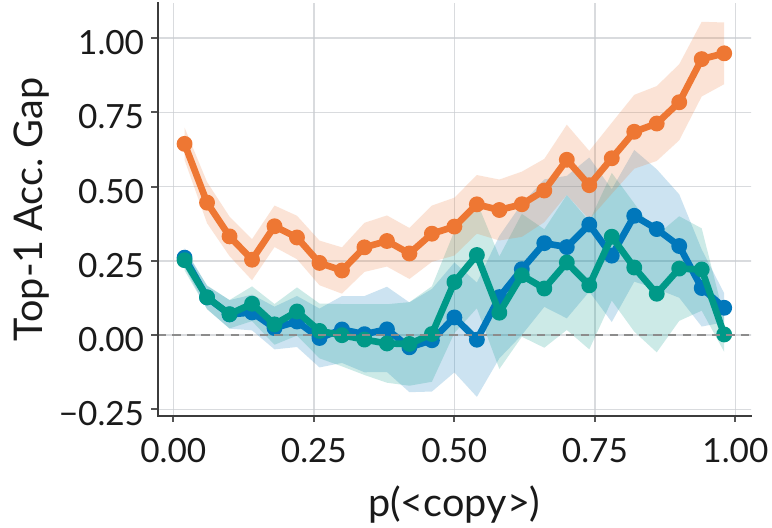}
    \caption{Top-1 accuracy gap}\label{fig:copy_bins:top1}
  \end{subfigure}\hfill
  \begin{subfigure}[t]{0.32\linewidth}
    \includegraphics[width=\linewidth]{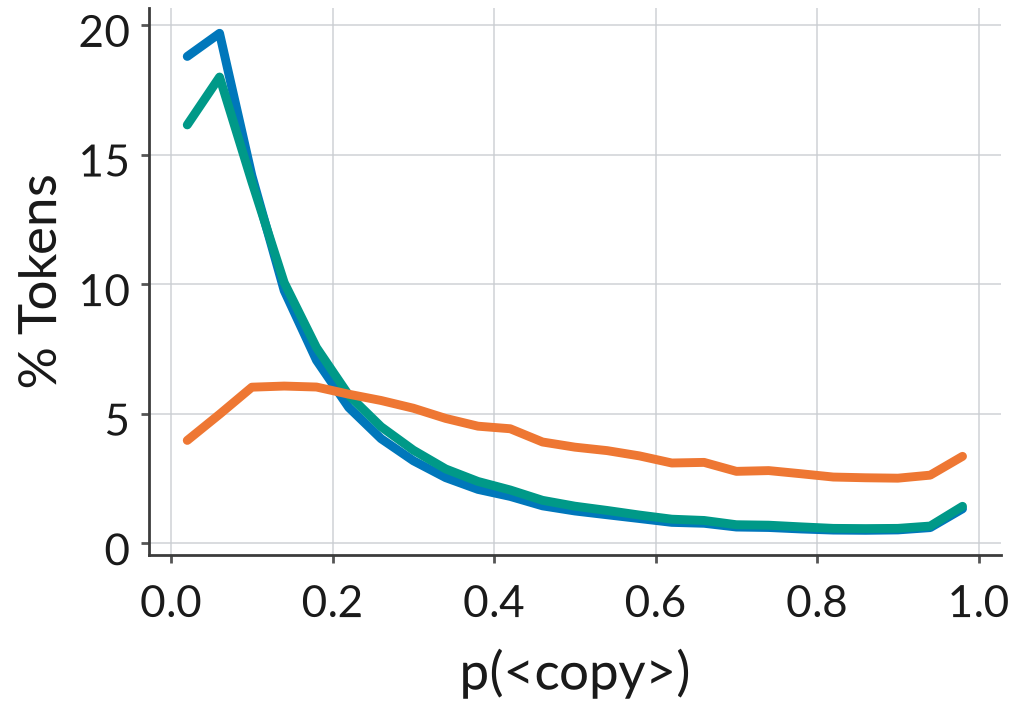}
    \caption{Distribution of $c_t$}\label{fig:copy_bins:dist}
  \end{subfigure}
  \caption{Prediction quality as a function of the copy gate $c_t = p(\texttt{<copy>})$ of our 1.7B CPLMs, on the FineWeb-Edu validation set. Shaded areas show the standard deviation within each bin. \textbf{(a)} Mean NLL difference between each CPLM and its vanilla counterpart, computed as CPLM minus vanilla (negative values favor CPLMs) and \textbf{(b)} top-1 accuracy gap (positive values favor CPLMs), computed on the tokens of each bin of $c_t$. \textbf{(c)} Share of tokens in each bin.}
  \label{fig:copy_bins}
\end{figure*}

We now study whether the gains of CPLMs are limited to the tokens that they copy. We bin tokens according to the probability $c_t$ that the CPLM assigns to \texttt{<copy>}, and compare the NLL and top-1 accuracy of CPLMs and vanilla models within each bin (\Cref{fig:copy_bins}). In every bin and for every backbone, CPLMs reach a lower NLL than their vanilla counterparts (\Cref{fig:copy_bins:nll}). This includes tokens for which $c_t$ is close to 0, where the prediction relies almost entirely on the LM head: the CP head also improves the predictions of the LM head itself, which is consistent with a better-trained backbone. The gains mostly grow with $c_t$, as the CP head contributes more to the prediction, and are largest for the SW CPLM, which cannot rely on its backbone to reach tokens beyond its window. Top-1 accuracy gains are also largest for tokens that are likely to be copied, especially for the SW CPLM, while Full and Hybrid CPLMs are on par with their vanilla counterparts for intermediate values of $c_t$ (\Cref{fig:copy_bins:top1}). Finally, Full and Hybrid CPLMs assign a low copy probability to most tokens, while the SW CPLM uses the CP head more uniformly (\Cref{fig:copy_bins:dist}).

\subsection{Copy Distance}\label{app:analysis:copy_dist}

\begin{figure}[ht]
  \centering
  \includegraphics[height=1.5em]{imgs/val_nll_curves_legend.pdf}\\[2pt]
  \includegraphics[width=0.4\linewidth]{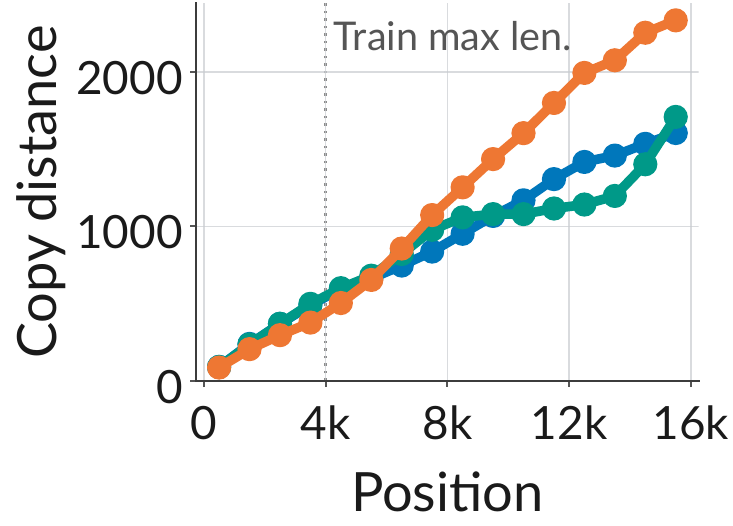}
  \caption{Average distance between the current position and the positions copied by the CP head, as a function of the position, for our 1.7B CPLMs on 5,000 FinePDFs documents of 16k tokens.}
  \label{fig:rolling:dist}
\end{figure}

\Cref{fig:rolling:dist} reports the average distance between the current position and the positions copied by the CP head, on the same FinePDFs documents as \Cref{fig:rolling}. For all backbones, the copy distance grows almost linearly with position, including beyond the 4k-token pretraining context: CPLMs do not only copy from the recent context, but make use of the whole document. The SW CPLM copies from the furthest positions, more than 2,000 tokens back at 16k tokens, even though its backbone only attends to the last 256 tokens.

\subsection{Other Languages}\label{app:analysis:lang}

\begin{figure}[ht]
  \centering
  \includegraphics[width=0.55\linewidth]{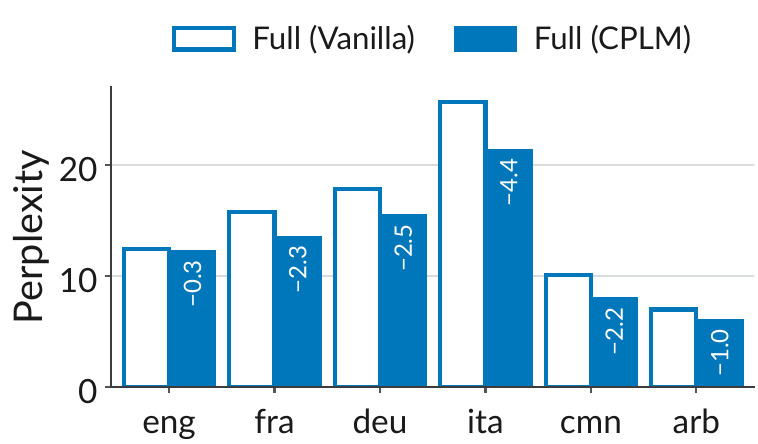}
  \caption{Perplexity of our 135M full-attention vanilla model and CPLM, trained on English FineWeb-Edu, on English data from FineWeb~\citep{fineweb} and on French, German, Italian, Mandarin Chinese and Arabic data from FineWeb2~\citep{penedo2025fineweb2}. Numbers indicate the perplexity difference between the CPLM and the vanilla model.}
  \label{fig:lang_ppl}
\end{figure}

Our models are pretrained on English data only. \Cref{fig:lang_ppl} evaluates our 135M full-attention models on other languages, which are largely out of their training distribution. On English FineWeb, the CPLM improves perplexity by 0.3 points, i.e. about 2\% in relative terms. On every other language, the gains are much larger, ranging from 14\% (Arabic) to 22\% (Mandarin Chinese) in relative terms, and reaching 4.4 points for Italian. This is in line with the larger gains observed on out-of-distribution English data (\Cref{fig:ppl_vs_memory}): when the model is less able to predict the next token from its parametric knowledge, copying tokens from the context becomes relatively more useful.

\clearpage
\section{Where Does the Gain Come From?}\label{app:mechanism}

In \Cref{sec:analysis:softmax}, we argue that CPLMs work mostly because the CP head provides a gradient pathway that bypasses the LM head bottleneck. This section tests this argument mechanically, by removing each ingredient in turn. First, we keep the CP head as an output layer but remove its gradient benefits, by stopping its gradient into the backbone (\Cref{app:mechanism:stopgrad}). Second, we keep the full CPLM but remove the LM head bottleneck itself, by training byte-level models whose vocabulary is smaller than their hidden dimension (\Cref{app:mechanism:bytes}). If our argument holds, both interventions should remove most of the gains of CPLMs.

All runs in this section use the 135M full-attention backbone and the training recipe of \Cref{tab:pretrain-hparams} (19{,}000 steps of 512 sequences of 2{,}048 units), with a single seed. As in \Cref{fig:rolling}, we report the mean copy probability $c_t$ and the mean sink probability $s_t$ over validation positions.

\subsection{Stopping the Copy Gradient into the Backbone}\label{app:mechanism:stopgrad}

\paragraph{Setup.} In a CPLM, the next-token loss reaches the backbone through two paths: the LM head, which also produces the copy gate $c_t$ through the \texttt{<copy>} token, and the CP head, through the queries $q_t = W_Q h_t$, the keys $k_\tau = W_K h_\tau$, and the sink score. We only cut the second path, by computing the queries and keys from a detached copy of the hidden states:
\begin{equation*}
    q_t = W_Q \operatorname{sg}(h_t), \qquad k_\tau = W_K \operatorname{sg}(h_\tau),
\end{equation*}
where $\operatorname{sg}$ is the stop-gradient operator. $W_Q$, $W_K$ and the sink key are still trained with the full mixture loss of \Cref{eq:proba_sink}, but they must now work with the representations that the backbone learns for the LM head. Everything else is identical to the CPLM. We verified that, with the stop-gradient in place, a loss computed from the CP head distribution alone sends no gradient into the backbone.

\begin{figure*}[ht!]
  \centering
  \includegraphics[height=1.5em]{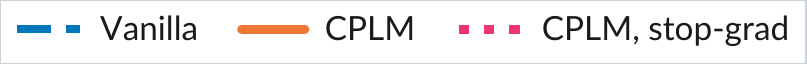}\\[2pt]
  \begin{subfigure}[t]{0.33\linewidth}
    \includegraphics[width=\linewidth]{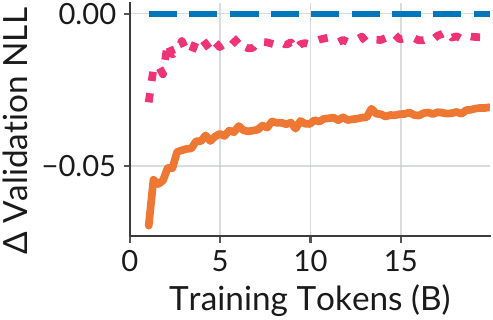}
    \caption{Validation NLL gap}\label{fig:stopgrad:delta}
  \end{subfigure}\hfill
  \begin{subfigure}[t]{0.33\linewidth}
    \includegraphics[width=\linewidth]{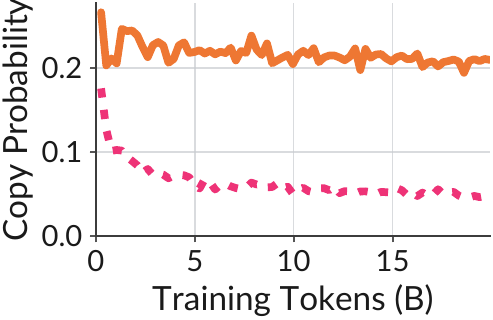}
    \caption{Copy probability $c_t$}\label{fig:stopgrad:copy}
  \end{subfigure}\hfill
  \begin{subfigure}[t]{0.33\linewidth}
    \includegraphics[width=\linewidth]{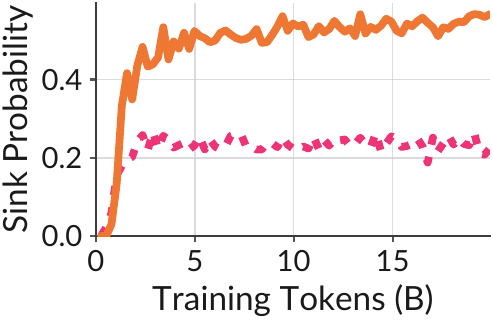}
    \caption{Sink probability $s_t$}\label{fig:stopgrad:sink}
  \end{subfigure}
  \caption{Stopping the gradient of the CP head into the backbone (135M, full attention, 20B tokens). \textbf{(a)} Validation NLL relative to the vanilla model at the same step. \textbf{(b--c)} Mean copy and sink probabilities. With the gradient stopped, the CPLM keeps about a quarter of its improvement over the vanilla model and copies much less.}
  \label{fig:stopgrad}
\end{figure*}

\begin{table}[ht!]
\centering
\small
\begin{tabular}{lccccc}
\toprule
 & Val.\ NLL & PPL & $\Delta$NLL vs.\ Vanilla & $c_t$ & $s_t$ \\
\midrule
Vanilla                 & 2.794 & 16.34 & ---      & ---   & ---   \\
CPLM                    & 2.763 & 15.85 & $-0.031$ & 0.210 & 0.567 \\
CPLM, stop-gradient     & 2.786 & 16.22 & $-0.008$ & 0.046 & 0.224 \\
\bottomrule
\end{tabular}
\caption{Final validation NLL (nats per token) and perplexity of the stop-gradient ablation, with the mean copy and sink probabilities.}
\label{tab:stopgrad}
\end{table}

\paragraph{Results.} Stopping the gradient removes most of the benefit (\Cref{fig:stopgrad,tab:stopgrad}). The CPLM improves on the vanilla model by 0.031 nats per token (0.49 PPL), against 0.008 nats (0.12 PPL) with the stop-gradient, i.e. about 25\% of the full effect. This remaining gain is still several times larger than the seed-to-seed standard deviation of our main arms (0.015--0.037 PPL, \Cref{app:seeds}): the CP head alone, reading representations that were never shaped for it, gives a small but real improvement.

The dynamics of the gap are informative (\Cref{fig:stopgrad:delta}). Early in training, the stop-gradient model is well ahead of the vanilla model ($-0.029$ nats at 1B tokens): the CP head can already exploit easy repetitions that a young backbone represents crudely. This advantage shrinks to about $-0.010$ nats by 5B tokens and to $-0.008$ nats at the end of training. The gap of the full CPLM starts larger ($-0.070$), narrows to $-0.040$ at 5B and $-0.035$ at 10B tokens, and ends at $-0.031$ nats. The stop-gradient model also uses the CP head much less (\Cref{fig:stopgrad:copy,fig:stopgrad:sink}). Its copy probability falls from 0.18 after 250 steps to 0.046, against a stable 0.21 for the CPLM, so that its effective gate $\lambda_t = c_t(1 - s_t)$ is 0.036, against 0.091 for the CPLM. Its sink probability is also lower (0.22 vs.\ 0.57), as less copy mass needs to be routed back to the LM head. The model learns that a CP head that it cannot shape is worth little, and mostly turns it off.

\paragraph{Interpretation.} Most of the gain of CPLMs comes from what the CP head gradient does to the backbone, rather than from the CP head as an additional output layer. CPLMs have no auxiliary loss: the backbone only receives the gradient of the next-token loss, but through two paths instead of one. The path through the CP head is much less compressed than the one through the LM head (\Cref{sec:analysis:softmax}), so that the backbone receives a richer training signal, which in particular helps it represent the context features needed to copy, consistently with CPLM backbones needing fewer induction heads (\Cref{sec:analysis:induction}). With the gradient path closed, $W_Q$ and $W_K$ can only linearly re-read features learned for the LM head, which are a poor fit for position-level matching. This also explains why CPLMs improve the NLL of tokens that they do not copy (\Cref{app:additional:copy_nll}): the backbone they share with the LM head is better trained.

\subsection{Byte-Level Language Modeling}\label{app:mechanism:bytes}

\paragraph{Motivation.} If the CP head mainly helped by escaping the LM head bottleneck, its benefits should vanish when there is no bottleneck. Byte-level models provide exactly this setting. Their vocabulary has $V = 257$ entries (256 byte values and a document separator, plus \texttt{<copy>} for CPLMs), fewer than the hidden dimension $D = 576$. The LM head matrix $W \in \mathbb{R}^{V \times D}$ then has full row rank: every distribution over the vocabulary can be expressed, and no component of the logit gradient is lost to the nullspace of $W^\top$, unlike for the token-level LM head of \Cref{fig:bottleneck_analysis}. This setting is the limit of our SpamLang experiment (\Cref{sec:analysis:softmax}), where vanilla models only failed once the vocabulary became large compared to the hidden dimension.

\paragraph{Setup.} We train three byte-level models on the same data stream as our token-level models: a vanilla model, a CPLM, and a CPLM without the sink. Each block of SmolLM2 tokens is decoded back to its raw UTF-8 bytes, which is lossless for byte-level BPE, and the first 2{,}049 bytes are kept as the training sequence, with byte $b$ mapped to id $b+1$ and the document separator to id 0. The backbone, sequence length (2{,}048, now in bytes), batch size, optimizer and schedule are unchanged, so that each model sees about 20B bytes, i.e. about 5B tokens worth of text. Only the embedding and LM head shrink, for about 107M parameters in total. Losses are in nats per byte.

\begin{figure*}[ht!]
  \centering
  \includegraphics[height=1.5em]{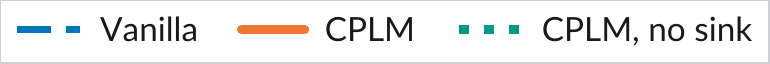}\\[2pt]
  \begin{subfigure}[t]{0.33\linewidth}
    \includegraphics[width=\linewidth]{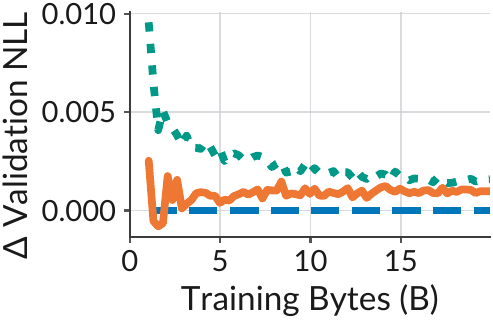}
    \caption{Validation NLL gap}\label{fig:bytes:delta}
  \end{subfigure}\hfill
  \begin{subfigure}[t]{0.33\linewidth}
    \includegraphics[width=\linewidth]{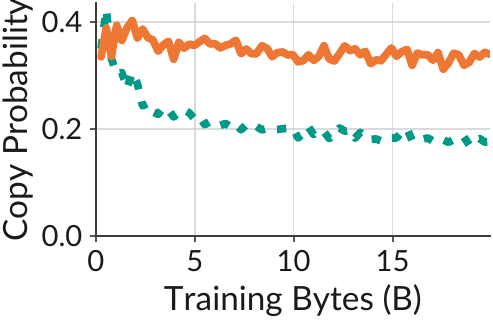}
    \caption{Copy probability $c_t$}\label{fig:bytes:copy}
  \end{subfigure}\hfill
  \begin{subfigure}[t]{0.33\linewidth}
    \includegraphics[width=\linewidth]{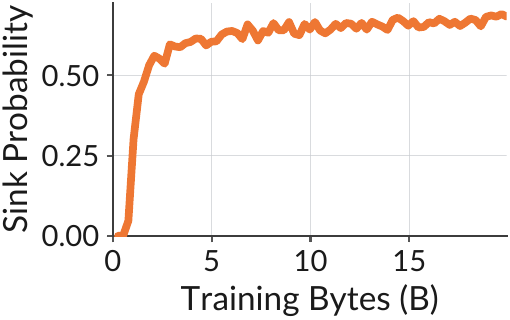}
    \caption{Sink probability $s_t$}\label{fig:bytes:sink}
  \end{subfigure}
  \caption{Byte-level language modeling (135M backbone, full attention, 20B bytes). \textbf{(a)} Validation NLL (nats per byte) relative to the vanilla model at the same step. \textbf{(b--c)} Mean copy and sink probabilities (sink for the CPLM with sink only). All three models reach the same loss within 0.002 nats per byte, even though the CPLM keeps a high copy probability throughout training.}
  \label{fig:bytes}
\end{figure*}

\begin{table}[ht!]
\centering
\small
\begin{tabular}{lcccc}
\toprule
 & Val.\ NLL (nats/byte) & Bits/byte & $c_t$ & $s_t$ \\
\midrule
Vanilla              & 0.6908 & 0.997 & ---   & ---   \\
CPLM                 & 0.6918 & 0.998 & 0.339 & 0.683 \\
CPLM, no sink        & 0.6924 & 0.999 & 0.176 & ---   \\
\bottomrule
\end{tabular}
\caption{Final validation loss of byte-level models, with the mean copy and sink probabilities.}
\label{tab:bytes}
\end{table}

\paragraph{Results.} At the byte level, the CP head brings no gain (\Cref{fig:bytes,tab:bytes}). Both CPLMs end within 0.002 nats per byte of the vanilla model ($+0.0010$ with the sink and $+0.0016$ without), which is comparable to the seed-to-seed standard deviation of our token-level models (\Cref{app:seeds}), and the training curves closely track each other throughout training. As for small backbone windows (\Cref{app:ablations:sink}), the sink helps early in training ($+0.0025$ vs.\ $+0.0096$ nats at 1B bytes), consistent with its role as a fallback while the CP head is still unreliable, but neither variant is ever meaningfully ahead of the vanilla model.

The CP head is not idle, however. The CPLM keeps a copy probability of about 0.34 throughout training, more than at the token level, and relies heavily on the sink ($s_t \approx 0.68$), which leaves an effective gate of 0.11, comparable to the 0.09 of the token-level CPLM. Without the sink, the copy probability settles at 0.18. In both cases, the model copies at least as much as at the token level, yet its loss does not improve: bytes repeat constantly, so copying is easy, but a full-rank LM head predicts them just as well.

\paragraph{Interpretation.} The byte-level results support the view that the token-level gains of CPLMs are tied to the LM head bottleneck. Once the LM head is full-rank and gradients reach the backbone without loss, pointing at the context gives the model nothing that it cannot express through the LM head. Together with the stop-gradient ablation, this draws a consistent picture: with a large vocabulary, the next-token loss reaches the backbone through the CP head as an additional training signal, bypassing the low-rank LM head (\Cref{sec:analysis:softmax}). This signal accounts for most of the gains of CPLMs, and it has nothing left to contribute once the LM head is no longer a bottleneck. Both interventions thus remove most of the gains of CPLMs, as predicted by our argument in \Cref{sec:analysis:softmax}.

\end{document}